\documentclass{article} % For LaTeX2e
\usepackage[final]{colm2026_conference}

\usepackage{microtype}
\usepackage{hyperref}
\usepackage{url}
\usepackage{booktabs}

\usepackage{lineno}

\usepackage{graphicx}
\usepackage[export]{adjustbox}
\usepackage{multirow}
\usepackage{tcolorbox}
\usepackage{comment}
\usepackage{amssymb}
\usepackage{ragged2e}
\usepackage{array}
\usepackage{amsmath}
\usepackage{xcolor}
\usepackage[table]{xcolor}
\usepackage[normalem]{ulem}

\definecolor{darkblue}{rgb}{0, 0, 0.5}
\hypersetup{colorlinks=true, citecolor=darkblue, linkcolor=darkblue, urlcolor=darkblue}

\title{The Percept-V Challenge: Can Multimodal LLMs Crack \\Simple Perception Problems?}

\author{Samrajnee Ghosh$^{1}$, Ashish Goswami$^{2}$, Naman Agarwal$^{1}$,  Hemanshu Garg$^{1}$,\\ 
\textbf{Chinmay Mittal$^{1}$}, \textbf{Mausam$^{1,2}$,} \textbf{Parag Singla$^{1,2}$}\\
$^{1}$Computer Science and Engineering, $^{2}$Yardi School of Artificial Intelligence\\
Indian Institute of Technology, Delhi\\
}

\begin{document}

\ifcolmsubmission
\linenumbers
\fi

\maketitle

\begin{abstract}
Cognitive science research treats visual perception, the ability to understand and make sense of a visual input, as one of the early developmental signs of intelligence. Its TVPS-4 framework categorizes and tests human perception into seven skills such as visual discrimination, and form constancy. Do Multimodal Large Language Models (MLLMs) match up to humans in basic perception? Even though many benchmarks evaluate MLLMs on advanced reasoning and knowledge skills, there is limited research that focuses evaluation on simple perception. In response, we introduce Percept-V, a dataset containing 6000 program-generated uncontaminated images divided into 30 domains, where each domain tests one or more TVPS-4 skills. Our focus is on perception, so we make our domains quite simple and the reasoning and knowledge required for solving them are minimal. Since modern-day MLLMs can solve much more complex tasks, our a-priori expectation is that they will solve these domains very easily. Contrary to our belief, our experiments show a weak performance of SoTA proprietary and open-source MLLMs compared to very high human performance on Percept-V. We find that as the number of objects in the image increases, performance goes down rather fast. Our experiments also identify the perception skills that are considerably harder for all models. Fine-tuning an open-source MLLM shows considerable gains in performance, though the gains only marginally carry over to other related datasets, pointing to limitation in generalization abilities of the learned representations. Code and data are available at \href{https://dair-iitd.github.io/Percept-V/}{https://dair-iitd.github.io/Percept-V/}
%ut our further analysis reveals gaps in the model's generalization abilities in light of visual perception.
\end{abstract}

\begin{comment}
\begin{figure*}[t]
    \centering
    \includegraphics[width=\linewidth]{images/Percept-V3.pdf}
    \caption{Sample questions from the Percept-V dataset illustrating tasks related to different visual perception skills as defined by the TVPS-4 framework. Each example shows the summarised image prompt, the correct answer, and a typical incorrect response from an MLLM.}
%     \label{fig:instances_part2}
\end{figure*}
\end{comment}

\section{Introduction}
\newcommand{\dataset}{Percept-V}
Multimodal Large Language Models (MLLMs) have achieved strong performance on complex tasks, prompting some claims that they are approaching human-level intelligence \citep{DBLP:conf/acl/ZhouXDY25, king2023administration} or even MENSA-level capabilities \citep{schregel2020intelligent}. However, existing visual benchmarks evaluate image understanding jointly with reasoning and domain-specific knowledge in areas such as engineering, medicine, and art \citep{nguyen2023brief, yue2024mmmu}. This integrated approach makes it challenging to isolate MLLMs' visual perception capabilities from their reasoning abilities and domain-specific knowledge.

\begin{figure}[t]
    \centering % Centers the content horizontally
    \begin{tikzpicture}[scale=0.5, every node/.style={transform shape}]
    
    % ---------- Top row: 3 taller objects ----------
    % Object 1
    \node[draw=black, rounded corners=5mm, thick, inner sep=5mm] at (-18,10) {
        \begin{minipage}[t][14cm]{7.5cm}
            \begin{center}{\bfseries Visual Closure}\end{center}
            \vspace{1mm}
            \begin{tcolorbox}[colback=yellow!40, colframe=yellow!50!black,
            width=\linewidth, boxrule=0.5pt, arc=3mm, left=1mm, right=1mm, top=1mm, bottom=1mm]
            Determine the number of shapes in the rows of the first column that match the outlines in the corresponding rows in the second column.
            \end{tcolorbox}
            \vspace{3mm}
            \begin{center}
                \includegraphics[scale = 0.4]{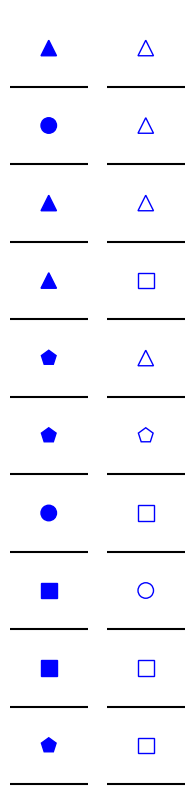}
            \end{center}
            \vspace{3mm}
            \begin{tcolorbox}[colback=green!40, colframe=green!50!black, width=\linewidth,
            boxrule=0.5pt, arc=3mm, halign=center]
            correct ans: 5
            \end{tcolorbox}
            \vspace{0.4mm}
            \begin{tcolorbox}[colback=red!40, colframe=red!70!black, width=\linewidth,
            boxrule=0.5pt, arc=3mm, halign=center]
            GPT-4o ans: ``ANSWER: 3''
            \end{tcolorbox}
        \end{minipage}
    };
    
    % Object 2
    \node[draw=black, rounded corners=5mm, thick, inner sep=5mm] at (-8.5,10) {
        \begin{minipage}[t][14cm]{7.5cm}
            \begin{center}{\bfseries Visual Sequential Memory}\end{center}
            \vspace{1mm}
            \begin{tcolorbox}[colback=yellow!40, colframe=yellow!50!black,
            width=\linewidth, boxrule=0.5pt, arc=3mm, left=1mm, right=1mm, top=1mm, bottom=1mm]
            This image contains some concentric circles in different colors. Write down the sequence of colors seen from  inside to outside.
            \end{tcolorbox}
            \vspace{3mm}
            \begin{center}
                \includegraphics[scale = 0.5]{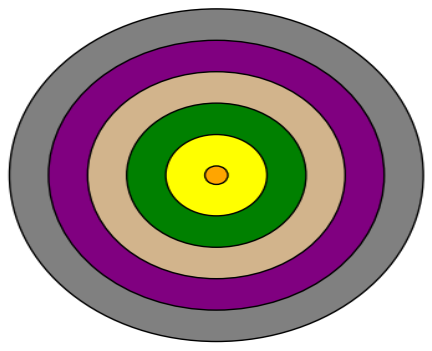}
            \end{center}
            \vspace{3mm}
            \begin{tcolorbox}[colback=green!40, colframe=green!50!black, width=\linewidth,
            boxrule=0.5pt, arc=3mm, halign=center]
            correct ans: orange, yellow, green, tan, purple, gray
            \end{tcolorbox}
            \vspace{0.5mm}
            \begin{tcolorbox}[colback=green!40, colframe=red!70!black, width=\linewidth,
            boxrule=0.5pt, arc=3mm, halign=center]
            GPT-4o ans: orange, yellow, green, tan, purple, gray
            \end{tcolorbox}
        \end{minipage}
    };
    
    % Object 3
    \node[draw=black, rounded corners=5mm, thick, inner sep=5mm] at (1,10) {
        \begin{minipage}[t][14cm]{7.5cm}
            \begin{center}{\bfseries Visual Discrimination}\end{center}
            \vspace{1mm}
            \begin{tcolorbox}[colback=yellow!40, colframe=yellow!50!black,
            width=\linewidth, boxrule=0.5pt, arc=3mm, left=1mm, right=1mm, top=1mm, bottom=1mm]
            The images are identical except for the fact that the lower image has a few objects missing. Count the number of missing objects in the lower image.
            
            \end{tcolorbox}
            %\vspace{8mm}
            \begin{center}
                \includegraphics[scale = 0.15]{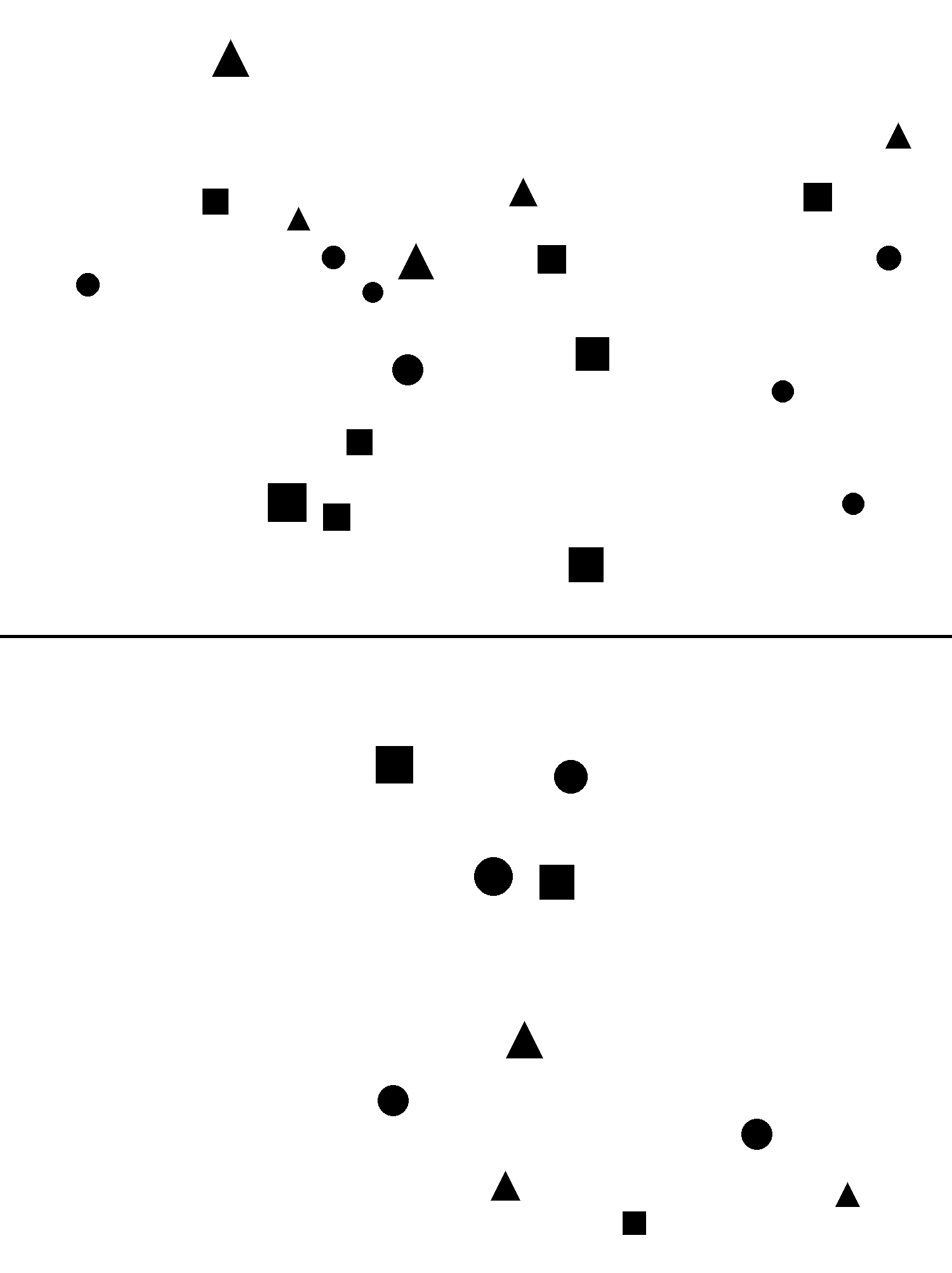}
            \end{center}
            %\vspace{8mm}
            \begin{tcolorbox}[colback=green!40, colframe=green!50!black, width=\linewidth,
            boxrule=0.5pt, arc=3mm, halign=center]
            correct ans: 10
            \end{tcolorbox}
            \vspace{0.5mm}
            \begin{tcolorbox}[colback=red!40, colframe=red!70!black, width=\linewidth,
            boxrule=0.5pt, arc=3mm, halign=center]
            GPT-4o ans: ``COUNT: 5''
            \end{tcolorbox}
        \end{minipage}
    };
    
    % ---------- Bottom row: 4 shorter objects ----------
    % Object 4
    \node[draw=black, rounded corners=5mm, thick, inner sep=5mm] at (-19,-5) {
        \begin{minipage}[t][12cm]{5.6cm}
            \begin{center}{\bfseries Visual Figure Ground }\end{center}
            \vspace{1mm}
            \begin{tcolorbox}[colback=yellow!40, colframe=yellow!50!black,
            width=\linewidth, boxrule=0.5pt, arc=3mm, left=1mm, right=1mm, top=1mm, bottom=1mm]
            Identify the shapes drawn over the background image. Form a set of all the unique shapes seen in the image.
            \end{tcolorbox}
            \vspace{1mm}
            \begin{center}
            \includegraphics[scale=0.5]{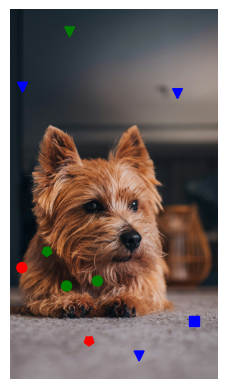}
            \end{center}
            \vspace{1mm}
            \begin{tcolorbox}[colback=green!40, colframe=green!50!black, width=\linewidth,
            boxrule=0.5pt, arc=3mm, halign=center]
            correct ans: circle,pentagon,rectangle,triangle
            \end{tcolorbox}
            \vspace{0.5mm}
            \begin{tcolorbox}[colback=red!40, colframe=red!70!black, width=\linewidth,
            boxrule=0.5pt, arc=3mm, halign=center]
            GPT-4o ans: ``{circle,triangle}''
            \end{tcolorbox}
        \end{minipage}
    };
    
    % Object 5
    \node[draw=black, rounded corners=5mm, thick, inner sep=5mm] at (-12,-5) {
        \begin{minipage}[t][12cm]{5.6cm}
            \begin{center}{\bfseries Visual Memory}\end{center}
            \vspace{1mm}
            \begin{tcolorbox}[colback=yellow!40, colframe=yellow!50!black,
            width=\linewidth, boxrule=0.5pt, arc=3mm, left=1mm, right=1mm, top=1mm, bottom=1mm]
            This is an image with some shapes drawn and numbers written on top of the shapes. Determine the list of numbers written on top of the circles.
            \end{tcolorbox}
            \vspace{8mm}
            \begin{center}
              \includegraphics[scale=0.4]{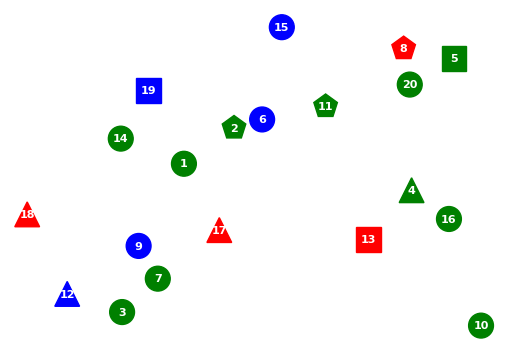}
            \end{center}
            \vspace{8mm}
            \begin{tcolorbox}[colback=green!40, colframe=green!50!black, width=\linewidth,
            boxrule=0.5pt, arc=3mm, halign=center]
            correct ans: 1,3,6,7,9,10,14,15,16,20
            \end{tcolorbox}
            \vspace{0.5mm}
            \begin{tcolorbox}[colback=red!40, colframe=red!70!black, width=\linewidth,
            boxrule=0.5pt, arc=3mm, halign=center]
            GPT-4o ans: ``CIRCLES:1,3,7,10,14,15''
            \end{tcolorbox}
        \end{minipage}
    };
    
    % Object 6
    \node[draw=black, rounded corners=5mm, thick, inner sep=5mm] at (-5,-5) {
        \begin{minipage}[t][12cm]{5.6cm}
            \begin{center}{\bfseries Visual Spatial Relationship}\end{center}
            \vspace{1mm}
            \begin{tcolorbox}[colback=yellow!40, colframe=yellow!50!black,
            width=\linewidth, boxrule=0.5pt, arc=3mm, left=1mm, right=1mm, top=1mm, bottom=1mm]
            Count the number of circles above the rectangle and below the rectangle
            \end{tcolorbox}
            \vspace{10mm}
            \begin{center}
            \includegraphics[scale=0.3]{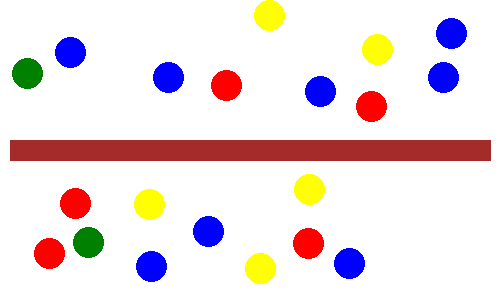}
            \end{center}
            \vspace{10mm}
            \begin{tcolorbox}[colback=green!40, colframe=green!50!black, width=\linewidth,
            boxrule=0.5pt, arc=3mm, halign=center]
            correct ans: ``ABOVE:10,BELOW:10''
            \end{tcolorbox}
            \vspace{0.5mm}
            \begin{tcolorbox}[colback=red!40, colframe=red!70!black, width=\linewidth,
            boxrule=0.5pt, arc=3mm, halign=center]
            GPT-4o ans: ``ABOVE:8,BELOW:10''
            \end{tcolorbox}
        \end{minipage}
    };
    
    % Object 7
    \node[draw=black, rounded corners=5mm, thick, inner sep=5mm] at (2,-5) {
        \begin{minipage}[t][12cm]{5.6cm}
            \begin{center}{\bfseries Visual Form Constancy}\end{center}
            \vspace{1mm}
            \begin{tcolorbox}[colback=yellow!40, colframe=yellow!50!black,
            width=\linewidth, boxrule=0.5pt, arc=3mm, left=1mm, right=1mm, top=1mm, bottom=1mm]
            Count the total number of circles in the image.
            \end{tcolorbox}
            \vspace{10mm}
            \begin{center}
                \includegraphics[scale=0.31]{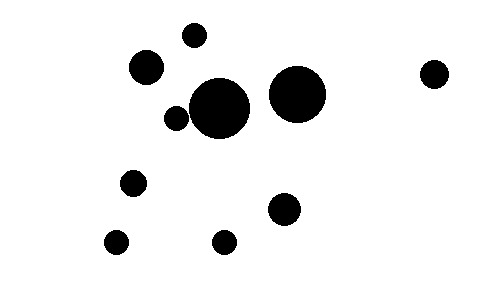}
            \end{center}
            \vspace{10mm}
            \begin{tcolorbox}[colback=green!40, colframe=green!50!black, width=\linewidth,
            boxrule=0.5pt, arc=3mm, halign=center]
            correct ans: 10
            \end{tcolorbox}
            \vspace{0.5mm}
            \begin{tcolorbox}[colback=green!40, colframe=red!70!black, width=\linewidth,
            boxrule=0.5pt, arc=3mm, halign=center]
            GPT-4o ans: ``CIRCLES:10''
            \end{tcolorbox}
        \end{minipage}
    };
    
    \end{tikzpicture}
    \caption{Sample questions from \dataset{} illustrating tasks related to different visual perception skills as defined by the TVPS-4 framework. Each example shows the summarised image prompt, the correct answer, and a typical incorrect response from an MLLM.}
    \label{fig:instances}
    %\label{fig:my_tikz_figure}
\end{figure}

Visual perception is a foundational cognitive ability that precedes complex reasoning and knowledge-based skills in humans \citep{Rabindran2020PiagetsTA}. Recent research has raised concerns about MLLMs' perceptual capabilities \citep{DBLP:journals/corr/abs-2402-07384, DBLP:conf/eccv/FuHLFWLRSMK24}, highlighting the need for a dedicated perception benchmark. To address this, we introduce Percept-V, a dataset of $6000$ programmatically generated, uncontaminated image-based problems spanning $30$ domains. These tasks primarily focus on visual understanding while requiring minimal reasoning and no specialized knowledge (see Figure \ref{fig:instances}). Our dataset is grounded in cognitive science, adopting the Test of Visual Perceptual Skills (TVPS-4) framework \citep{martin2006test}, which categorizes visual perception into $7$ skills, including visual discrimination, and spatial relationships (Section \ref{s:dataset}). Each domain evaluates 1–3 of these skills using a standardized question prompt describing the task. Instances involve a basic perception question over an image with simple geometric objects (e.g., circles, triangles) varying in colour and size. Such synthetic 2D images have applications in robotics \citep{li2024mmromultimodalllmseligible} and mathematics \citep{wang2024measuring}.

To systematically assess performance across various difficulty levels, each domain contains 200 images with varying problem size, which is operationalised through object count, grid size, or required processing steps. This enables us to evaluate MLLM performance as a function of visual complexity and provides insight into how they handle increasing perceptual demands.
We evaluate 4 SOTA proprietary MLLMs (GPT-5-mini, GPT-4o, o4-mini, Gemini 2.5 Flash), 4 open-source models (Qwen 2.5 VL Instruct-7B, DeepSeek VL2 Tiny, Qwen-3VL-8B-Thinking and  Gemma-4-31B-it), and also conduct a human study.  Our findings show significant gaps between human and model performance, with consistent degradation across all evaluated models as problem size increases, suggesting systematic limitations in handling visual complexity. While some models perform relatively better on specific skills, overall performance spans a wide range across models, indicating broad visual perceptual limitations. We also investigate the effects of LoRA and full finetuning on our dataset. Overall, our work is an important step towards understanding the visual perceptual limitations that must be addressed in the future to develop more robust MLLMs.

\begin{table*}[h!]

\centering
\renewcommand{\arraystretch}{1.5}
\setlength{\tabcolsep}{6pt}
\begin{tabular}{|p{3cm}|p{10.1cm}|}
\hline
{\small\textbf{Skill}} & {\small\textbf{Domain}} \\
\hline
{\small\RaggedRight\hyphenpenalty=10000 Visual Discrimination (D)} & {\small change colour, vanishing objects, comparing size, counting shapes, sort lines, sort circles, locate circles colour, locate circles shape, match shadow, match outline} \\ \hline
{\small\RaggedRight\hyphenpenalty=10000 Visual Memory (M)} & {\small list colours, list shapes, circle boxes, colours present, count coloured circles, numbered shapes, graph counting, sort circles, sort lines, counting shapes, identifying shapes} \\ \hline
{\small\RaggedRight\hyphenpenalty=10000 Visual Sequential Memory (SM)} & {\small grid path, layered colours, layered shapes, comparing size, match outline, match shadow} \\ \hline
{\small\RaggedRight\hyphenpenalty=10000 Visual Figure Ground (FG)} & {\small list colours, list shapes, locate circles colour, locate circles shape, layered shapes, layered colours} \\ \hline
{\small\RaggedRight\hyphenpenalty=10000 Visual Form Constancy (FC)} & {\small mirror image, water image, counting circles, counting shapes, numbered shapes} \\ \hline
{\small\RaggedRight\hyphenpenalty=10000 Visual Closure (C)} & {\small match outline, match shadow, layered shapes, layered colours} \\ \hline
{\small\RaggedRight\hyphenpenalty=10000 Visual Spatial Relationship (SR)} & {\small comparing size, circle location, circle right triangle, counting locations, cross and knots, inside circles, maze solving} \\ \hline
\end{tabular}
\caption{Domains using each skill. Note that a single domain can appear in multiple rows in the above table since multiple skills may be required to solve its Problem Instances}
\label{table:skills_domain}
\end{table*}

\section{Related Work}
A large number of benchmarks exist for text-only problems, and modern-day LLMs have shown excellent performance in complex reasoning tasks \citep{fcorebench, jeebench}.
Most existing benchmarks for evaluating MLLMs require a combination of perception, reasoning, and domain knowledge. For instance, NTSEBench \citep{pandya2024ntsebench} evaluates commonsense reasoning, MathVista \citep{lu2023mathvista} focuses on compositional reasoning, and MathVision \citep{wang2024measuring} spans 16 mathematical disciplines. Similarly, OlympiadBench \citep{he2024olympiadbench} includes advanced problems from mathematics, and science, and IQBench \citep{pham2025iqbenchsmartvisionlanguagemodels} tests MLLMs on visually-centric IQ tests. Large-scale benchmarks such as MMMU, CMMMU, and EMMA \citep{yue2024mmmu,zhang2024cmmmu,DBLP:conf/icml/HaoGWLYW025} cover fields such as engineering, medicine, humanities, and business, while SciBench \citep{wang2023scibench} contains college-level science questions. Datasets like ManBench and Rebus \citep{DBLP:conf/acl/ZhouXDY25, das2025leftcirclearrowrightboxedtextbusrightlargediversemultimodal} require complex reasoning over images, whereas other datasets require interpreting abstract figures and plots \citep{li2024multimodal}. All these benchmarks entangle perception with reasoning and sometimes specialized knowledge.

Recent work has attempted to disentangle these abilities. MatHLENS \citep{chung2025mllmslearnlearnmultimodal} separately evaluates perception, reasoning, and integrating abilities. Other studies highlight perceptual limitations of MLLMs \citep{lee2025exploringmultimodalperceptionlarge}, including poor performance on small or low-quality objects and fine-grained perception \citep{yan2025visuriddlesfinegrainedperceptionprimary}, and introduce simple perception-focused benchmarks \citep{DBLP:conf/eccv/FuHLFWLRSMK24}. Benchmarks like VOILA \citep{DBLP:conf/iclr/YilmazPLGBJY25} test perceptual understanding through image generation. However, these works typically rely on complex scenes and do not analyze the impact of problem size. 

Closest to our work are contemporary benchmarks like \citet{DBLP:conf/eacl/KanadeG26, huang2026humancognitivebenchmarksreveal, cao2025visualcognitiongaphumans, kamoi2025visonlyqalargevisionlanguage} that focus on perception without any specialized knowledge. Compared to these papers, our work introduces a plug-and-play pipeline for generating a dataset of any size with only simple instructions. Gold answers are also generated automatically, eliminating the need for human annotation. In contrast to \citet{DBLP:conf/eacl/KanadeG26}, our dataset provides 30 novel domains grounded in all 7 skills of the TVPS-4 framework versus their 12 and the variation in problem size in our dataset is much more fine-grained which allows for systematic scaling in task complexity. Unlike VISFACTOR \citep{huang2026humancognitivebenchmarksreveal} which extracts gold truth answers from FRCT battery and lacks a completely automated evaluation process, both our gold-answer generation and evaluation of model performance are fully done programmatically. We also provide detailed analysis of skill-specific weaknesses, scaling behaviour, answer-type difficulties and post-finetuning generalization gaps in MLLMs. Unlike VisOnlyQA which uses a mix of synthetic and real images, Percept-V uses purely synthetic images with controllable parameters and focuses specifically on basic visual perception, enabling systematic evaluation without the influence of reasoning, semantic understanding, or domain knowledge.

\section{Dataset}
\label{s:dataset}
We describe our dataset, {\dataset{}}, which consists of 30 domains designed to evaluate {\em basic perception} skills based on the TVPS-4 cognitive framework \citep{martin2006test}. TVPS-4, being a popular test of visual perception in humans, helps us to ground our dataset in well-established parameters for evaluating perception. TVPS-4 defines seven visual perceptual skills: (1) {\it Visual Discrimination} (identifying similarities/differences in size, color, shape), (2) {\it Visual Memory} (recalling visual traits), (3) {\it Visual Sequential Memory} (recalling ordered sequences), (4) {\it Visual Spatial Relationships} (understanding object relationships), (5) {\it Visual Figure Ground} (locating objects in clutter), (6) {\it Visual Form Constancy} (recognizing shapes under transformations), and (7) {\it Visual Closure} (recognizing incomplete forms).

\begin{comment}
\begin{enumerate}
    \setlength{\itemsep}{0pt}
    \item {\it Visual Discrimination}: The ability to determine differences or similarities in objects based on size, colour, shape, etc.
    \item {\it Visual Memory}: The ability to recall visual traits of a form or object.     
   \item {\it Visual Sequential Memory}: The ability to recall a sequence of objects in the correct order. 
 \item {\it Visual Spatial Relationships}: Understanding the relationships between objects. 
    \item {\it Visual Figure Ground}: The ability to locate something in a busy background.  
   \item {\it Visual Form Constancy}: The ability to know that a form or shape is the same, even if it has been made smaller/larger or has been rotated. 
   \item {\it Visual Closure}: The ability to recognize a form, when a part of the picture is missing. 
\end{enumerate}
\end{comment}
 
Each domain evaluates one or more of these skills and contains multiple {\em problem instances} with varying perceptual complexity. Overall, 17 domains test a single skill, 7 test two skills, and 6 test three skills jointly. Table~\ref{table:skills_domain} lists, for each skill, the domains that require that skill to solve the problem instances correctly. Looking further at the performance on the 17 single skill domains, we found that some skills are more difficult for most models than others. Fig \ref{fig:single_skill} in section \ref{sec:single_skill_appendix} shows that 'vsm' i.e., Visual Sequential Memory is a weakness for 4 out of 8 models as they exhibit their lowest performance in that skill.
Notably, visual memory and visual
sequential memory traditionally involve recalling information after it is removed from view. Since MLLMs retain all input in their memory, our setup only approximates these skills as scanning and transcription tasks (which are a subset to the original skill). Despite this simplification, model performance remains low (Table~\ref{table:main_results}), indicating fundamental perceptual limitations. Please refer to appendix section~\ref{sec:long_term_visual_memory} for a detailed discussion.

% \AG{We argue a faithful analogue survives: the vision encoder maps the image to latent embeddings before the question is seen, and the model answers by attending to that fixed, lossy encoding rather than to raw pixels mimicking the position of a human consulting a mental image. The context window is thus a transient short-term memory whose fidelity and capacity our tasks measure. Recall from long-term memory maps instead onto retrieval from the model's weights, which we test in section~\ref{sec:long_term_visual_memory}.}

Each domain is associated with a unique question prompt, and every problem instance is a pair $(Q, I)$, where $Q$ is the domain-specific question prompt and $I$ is a Python programme generated image.\footnote{Image generation is controlled by domain-specific hyperparameters (Appendix \ref{sec:appendix}).} The same prompt is applied across images within a domain. Instances vary in {\em problem size}, which captures perceptual complexity through factors such as number of objects, distractors, or required processing steps. Each domain includes 20 sizes varying from 1 to 20, with 10 instances per size, yielding 200 instances per domain and 6000 instances overall. For example, in the {\em counting\_circles} domain, size corresponds to the number of circles, while in {\em numbered\_shapes}, both target objects and distractors increase with size. Sample instances are shown in Figure~\ref{fig:instances}.

Each question prompt $Q$ consists of three components $(in, r, op)$: $in$ describes the input image, $r$ specifies the task in a step-by-step manner with detailed instructions, and $op$ defines the required answer format. The dataset includes four answer types : (i) single-value outputs (Boolean or numeric) (11 domains), (ii) fixed-length outputs (tuples or lists of constant size) (6 domains), (iii) ordered lists (sequence-sensitive outputs whose length varies with problem size) (8 domains), and (iv) unordered sets (collections of unique elements) (5 domains). Each prompt explicitly specifies the output format to enable automatic evaluation. In practice, Most MLLMs adhere well to formatting constraints; errors primarily arise from incorrect perception rather than format violations (Section \ref{sec:other}).

All images are generated using automated scripts -- one script per domain. Each script takes as input a set of problem sizes and the number of instances per size, and randomizes object placement to ensure diversity. Images consist of simple geometric shapes (e.g., circles, triangles, squares) arranged in basic structures such as grids, rows, and mazes. %, created using standard Python libraries including Pillow, Matplotlib, and OpenCV. 
Ground-truth answers are computed during generation, eliminating ambiguity or annotation errors. Our programmatic generation provides two key advantages: (1) precise control over problem complexity, enabling systematic evaluation across difficulty levels, and (2) elimination of dataset contamination, a known issue in existing benchmarks \citep{shojaee2025illusion}.

\section{Experiments}
Through our experiments, we aim to answer the following questions: (1) How well do state-of-the-art (SoTA) open and proprietary MLLMs perform on \dataset{}? (2) Do some models perform better than others, particularly for specific skills? (3) How does performance vary across skills — are some skills inherently harder? (4) How does performance change with increasing problem size? (5) How does MLLM performance compare to humans? (6) How does finetuning affect the performance on our dataset?

To answer these, we evaluate 4 proprietary MLLMs: GPT-4o, Gemini 2.5 Flash, o4-mini, and GPT-5-mini, and 4 open-weight models: Qwen-2.5-VL-7B-Instruct, DeepSeek-VL2-Tiny, Qwen-3-VL-8B-Thinking and Gemma-4-31B-it. Among these, GPT-5-mini, o4-mini, Gemini, Qwen3 and Gemma-4 support “think with image” capabilities \citep{su2025thinkingimagesmultimodalreasoning}, i.e., leverage visual information during intermediate reasoning.
Experiments for open-source models are conducted on an NVIDIA A100 GPU (details in Appendix \ref{sec:appendix}). To ensure reproducibility, hyperparameters such as temperature and top\_p are set to 0.0 wherever possible (restricted for some models). All experiments are conducted primarily in a zero-shot setting; one-shot and few-shot prompting were also tested but yielded negative results. For each domain, the same prompt (question, output format) is used across all models (see Appendix).

\begin{table*}[t!]
\centering
\small
\resizebox{\linewidth}{!}{
\begin{tabular}{|l|r|r|r|r|r|r|r|r||r|}
\hline
{\bf Skill / Domain} & {\bf GPT-5-mini} & {\bf GPT-4o} & {\bf o4-mini} & {\bf Gemini} & {\bf Gemma4} & {\bf Qwen3VL} & {\bf Qwen2.5VL} & {\bf DeepSeek} & {\bf Avg. Acc.}\\
\hline
\multicolumn{10}{|l|}{\bf --- Performance by Skill ---} \\ \hline

Visual Discrimination & 55.0 & 23.7 & 55.3 & 51.25 & 68.75 & 34.74 & 11.5 & 4.1 & 38.04 \\
Visual Memory & 49.95 & 33.55 & 47.55 & 46.55 & 64.0 & 30.12 & 19.59 & 1.32 & 36.58 \\
Visual Sequential Memory & 43.5 & 21.5 & 47.5 & 41.92 & 52.58 & 15.75 & 4.67 & 5.25 & 29.08 \\
Visual Figure Ground & 64.75 & 25.42 & 63.25 & 62.83 & 59.5 & 40.0 & 7.58 & 0.08 & 40.43 \\
Visual Form Constancy & 58.7 & 27.3 & 48.9 & 50.3 & 71.0 & 34.67 & 25.6 & 3.0 & 39.93 \\
Visual Closure & 44.88 & 23.0 & 50.25 & 43.25 & 43.5 & 13.12 & 5.75 & 7.88 & 28.96 \\
Visual Spatial Relationship & 66.64 & 50.36 & 62.64 & 69.79 & 84.5 & 54.36 & 27.79 & 17.64 & 54.22 \\
\hline
Average of skills & 54.77 & 29.26 & 53.63 & 52.27 & 63.4 & 31.82 & 14.64 & 5.61 & 38.18\\
\hline
\hline
\multicolumn{10}{|l|}{\bf --- Performance by Domain ---} \\ \hline
change\_colour & 32.0 & 18.5 & 32.0 & 23.5 & 58.5 & 8.5 & 8.0 & 4.5 & 23.19 \\
circle\_boxes & 30.0 & 33.0 & 34.0 & 28.0 & 64.5 & 15.0 & 12.5 & 10.5 & 28.44 \\
circle\_location & 56.5 & 41.5 & 46.0 & 57.5 & 81.5 & 22.5 & 14.5 & 5.5 & 40.69 \\
circle\_right\_triangle & 100.0 & 82.0 & 99.5 & 99.0 & 98.5 & 94.5 & 25.0 & 50.0 & 81.06 \\
colours\_present & 3.5 & 1.5 & 4.5 & 4.5 & 5.5 & 1.5 & 0.0 & 0.0 & 2.62 \\
comparing\_size & 27.0 & 36.5 & 31.0 & 55.0 & 81.0 & 31.5 & 5.0 & 0.0 & 33.38 \\
count\_coloured\_circles & 51.0 & 44.0 & 37.5 & 44.5 & 99.0 & 38.0 & 39.0 & 2.5 & 44.44 \\
counting\_circles & 68.0 & 57.5 & 55.0 & 55.5 & 100.0 & 54.5 & 42.5 & 5.5 & 54.81 \\
counting\_locations & 44.5 & 20.5 & 36.5 & 33.5 & 98.5 & 37.0 & 15.5 & 0.0 & 35.75 \\
counting\_shapes & 72.5 & 44.0 & 58.5 & 61.0 & 100.0 & 62.37 & 41.0 & 0.0 & 54.92 \\
cross\_and\_knots & 99.5 & 57.5 & 100.0 & 96.0 & 81.5 & 89.5 & 23.5 & 0.0 & 68.44 \\
graph\_counting & 39.0 & 31.0 & 33.5 & 36.0 & 51.5 & 25.5 & 26.0 & 0.0 & 30.31 \\
grid\_path & 54.5 & 0.5 & 53.0 & 23.5 & 60.5 & 10.5 & 0.0 & 0.0 & 25.31 \\
identifying\_shapes & 79.5 & 69.0 & 92.5 & 47.0 & 100.0 & 80.0 & 41.5 & 0.0 & 63.69 \\
inside\_circles & 96.0 & 87.5 & 95.5 & 93.0 & 97.0 & 94.5 & 96.5 & 68.0 & 91.0 \\
layered\_colours & 35.0 & 25.0 & 33.5 & 34.0 & 29.5 & 10.0 & 0.5 & 0.0 & 20.94 \\
layered\_shapes & 24.0 & 11.5 & 20.0 & 18.5 & 16.5 & 11.5 & 9.5 & 0.0 & 13.94 \\
list\_colours & 41.0 & 18.0 & 35.5 & 58.0 & 37.0 & 3.5 & 1.5 & 0.5 & 24.38 \\
list\_shapes & 88.5 & 70.5 & 90.5 & 92.0 & 90.5 & 31.5 & 21.0 & 0.0 & 60.56 \\
locate\_circles\_colour & 100.0 & 11.5 & 100.0 & 82.0 & 92.0 & 95.0 & 6.0 & 0.0 & 60.81 \\
locate\_circles\_shape & 100.0 & 16.0 & 100.0 & 92.5 & 91.5 & 88.5 & 7.0 & 0.0 & 61.94 \\
match\_outline & 66.0 & 23.0 & 84.5 & 72.5 & 82.0 & 19.0 & 7.5 & 17.0 & 46.44 \\
match\_shadow & 54.5 & 32.5 & 63.0 & 48.0 & 46.0 & 12.0 & 5.5 & 14.5 & 34.5 \\
maze\_solving & 43.0 & 27.0 & 30.0 & 54.5 & 53.5 & 11.0 & 14.5 & 0.0 & 29.19 \\
mirror\_image & 33.5 & 12.5 & 29.0 & 24.0 & 32.5 & 3.5 & 21.5 & 5.0 & 20.19 \\
numbered\_shapes & 81.5 & 12.0 & 75.5 & 88.5 & 84.0 & 49.5 & 8.0 & 1.0 & 50.0 \\
sort\_circles & 36.5 & 24.0 & 34.0 & 30.0 & 31.5 & 0.0 & 20.0 & 0.0 & 22.0 \\
sort\_lines & 26.5 & 22.0 & 27.0 & 22.5 & 40.5 & 24.5 & 5.0 & 0.0 & 21.0 \\
vanishing\_objects & 35.0 & 9.0 & 23.0 & 25.5 & 64.5 & 6.0 & 10.0 & 5.0 & 22.25 \\
water\_image & 38.0 & 10.5 & 26.5 & 22.5 & 38.5 & 3.5 & 15.0 & 3.5 & 19.75 \\
\hline
Full Percept-V Dataset & 55.22 & 31.65 & 52.7 & 50.75 & 66.92 & 34.48 & 18.1 & 6.43 & 39.53\\
\hline
\end{tabular}}
\caption{Comparison of MLLMs' performance across different skills and domains (in \%).}
\label{table:main_results}
\end{table*}

% \textcolor{red}{Update Score for colours\_present}

\paragraph{Domain-wise Analysis:} 
% \sout{Table~\ref{table:main_results} shows performance across domains and skills. Across most domains, MLLM performance is poor. Closed-source models outperform open-source ones, but still struggle significantly. Among proprietary models, GPT-4o (non-thinking) performs the worst, achieving $\le 20\%$ accuracy on over 12 of 30 domains and $\le 50\%$ on 24 domains. Reasoning models perform somewhat better: o4-mini achieves $\le 20\%$ on only 2 domains and $\le 50\%$ on 17, with Gemini and GPT-5-mini showing similar trends. Open-source models perform substantially worse: Qwen achieves $\le 20\%$ accuracy on 20 domains and exceeds 50\% on only one domain; DeepSeek shows comparable results. Overall, GPT-5-mini achieves the highest average performance, but only at 55.22\%.}

Table~\ref{table:main_results} shows performance across domains and skills. Across most
domains, MLLM performance is poor. Among proprietary models, GPT-4o (non-thinking) performs
the worst, achieving $\le 20\%$ accuracy on 11 of 30 domains and $\le 50\%$ on 24 domains.
Reasoning models perform somewhat better: o4-mini achieves $\le 20\%$ on only 2 domains and
$\le 50\%$ on 17, with Gemini and GPT-5-mini showing similar trends. The open-source models
span a far wider range than the proprietary ones: Gemma-4-31B is the strongest model in our
evaluation, falling $\le 50\%$ on only 9 domains and $\le 20\%$ on 2, and Qwen-3VL-8B already
edges out GPT-4o; at the other end Qwen-2.5VL is $\le 20\%$ on 20 domains and exceeds 50\% on
only one, while DeepSeek scores exactly zero on 16. Overall, Gemma-4-31B achieves
the highest average performance, but only at 66.92\%.

% The open/proprietary divide is therefore
% not the axis that separates strong from weak perception here.}

These results highlight that both proprietary and open-source MLLMs — regardless of their reasoning capability — fail to achieve strong performance across most domains. This contrasts sharply with human performance (Section \ref{subsubsec:human_study}). Additionally, performance patterns are consistent across models: domains such as {\em inside\_circles} show relatively higher performance across all models, whereas {\em colours\_present} consistently exhibits the lowest accuracy. The latter requires identifying colours of objects, and confusion between similar colours (e.g., turquoise vs. blue), despite provided swatches and legends, may explain the difficulty. This suggests that certain domains are intrinsically harder for current MLLMs.

\paragraph{Skill-wise Analysis:}
% \sout{Skill-level analysis reveals similar trends. GPT-5-mini achieves the highest average accuracy across skills (54.77\%), followed by o4-mini and Gemini. However, it exceeds 50\% accuracy on only 4 of 7 skills, with a maximum of 66.64\%, indicating limited robustness. GPT-4o performs significantly worse (28.6\%), while Qwen achieves only 14.48\%. We see that thinking models like GPT-5-mini, o4-mini and Gemini are consistently better than GPT-4o across skills, followed by open source models, which perform quite poorly.}

Skill-level analysis reveals similar trends. Gemma-4-31B achieves the highest average accuracy
across skills (63.40\%), followed by GPT-5-mini (54.77\%), o4-mini and Gemini. Even the best model achieves more than 70\% on only 2 of 7 skills while none of the other models cross this threshold, indicating sub-optimal performance across skills. GPT-4o performs
significantly worse (29.26\%), while Qwen-2.5VL achieves only 14.64\% and DeepSeek 5.61\%. A significant differentiator in model performance is whether it is thinking or not - every thinking model consistently outperforms the non-thinking open models which trail far behind.

Across all models, some skills are consistently easier or harder. Most models perform best on Visual Spatial Relationships, likely due to the prevalence of spatial descriptions (e.g., “left of”, “above”) in training data. In contrast, Visual Closure is among the hardest skills, involving tasks such as recognizing partially occluded shapes or matching outlines to filled objects — a phenomena likely underrepresented in training data. This consistent pattern suggests inherent skill-specific weaknesses in MLLMs.

\begin{figure*}[t!]
\includegraphics[width=\textwidth, left]{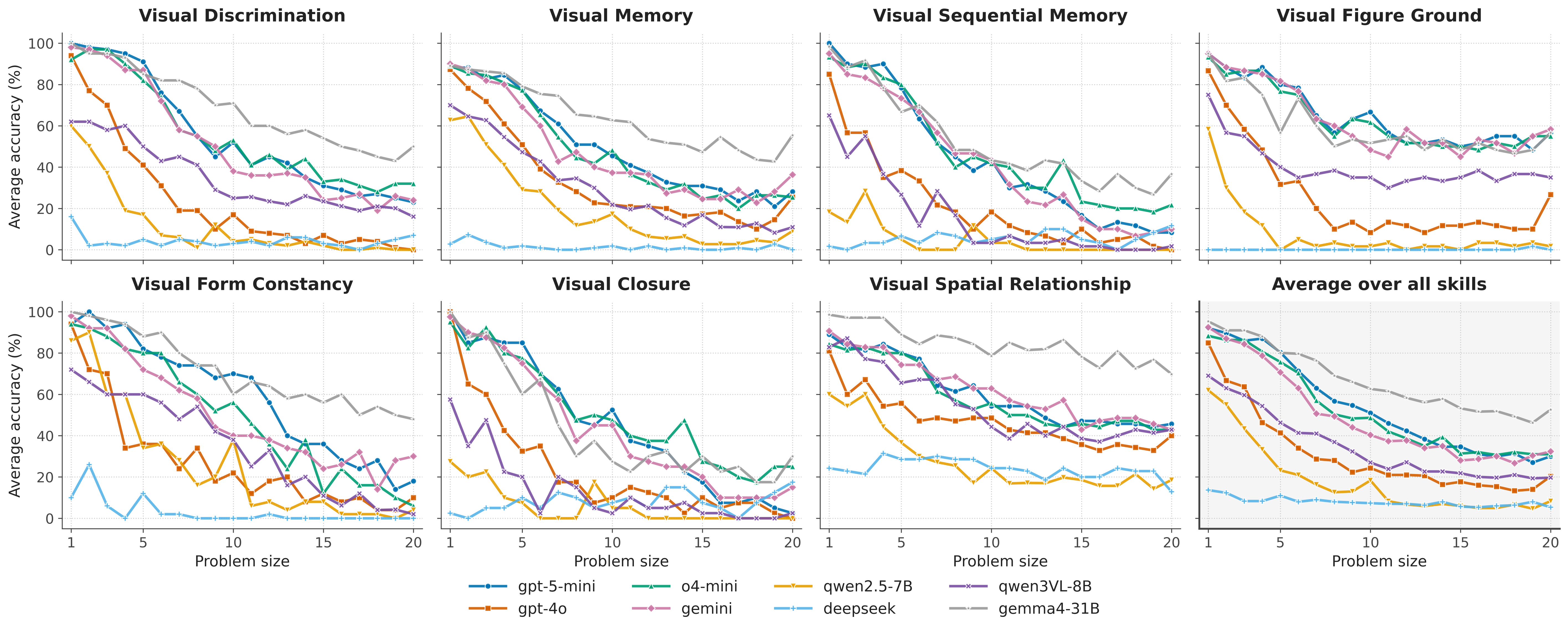}
\caption{The overall accuracy of all models in different skills.}
\label{fig:size_vs_acc}
\end{figure*}

\subsection{Performance vs Size} % should answer Q4.
To analyze performance scaling (Q4), we evaluate accuracy across varying problem sizes (Figure~\ref{fig:size_vs_acc}). Each of the seven plots corresponds to a skill, with an additional plot (last plot of second row) showing average performance across skills.

We observe a consistent and significant drop in performance as problem size increases across all skills and models. Reasoning models (Gemma-4, o4-mini, Gemini) often achieve near-perfect accuracy on smaller instances (fewer objects or simpler layouts), but performance degrades sharply as complexity increases. Gemma-4, GPT-5-mini, o4-mini, and Gemini are relatively more robust compared to others, though still exhibit substantial declines. GPT-4o follows, while other open-source models degrade most severely. These findings indicate that MLLMs lack robustness to increasing visual complexity and fail to generalize effectively across sizes.

\begin{figure*}[!t]
\centering
\includegraphics[width=\textwidth]{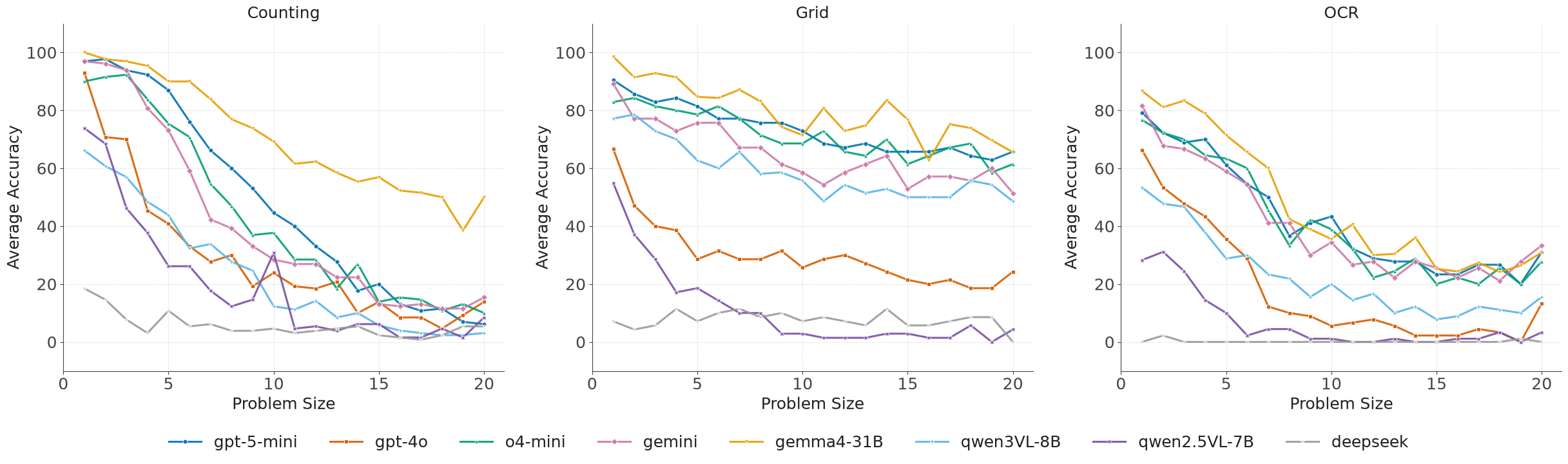}
\caption{The overall accuracy of all models in counting, grid understanding and ocr skills.}
\label{fig:alt_skills}
\end{figure*}

\subsection{Alternate Skill Analysis}
Beyond TVPS-4, three additional recurring skills emerge in our dataset: counting, optical character recognition (OCR), and grid understanding, appearing in 13, 9, and 7 domains respectively. While counting is a basic reasoning skill included mainly to facilitate automatic evaluation, it may distract from the main results. Comparing counting (C) vs. non-counting (NC) domains yields: GPT-4o (29.04 vs.\ 33.65),
GPT-5-mini (47.73 vs.\ 60.94), o4-mini (43.00 vs.\ 60.12), Gemini (40.92 vs.\ 58.26),
Gemma-4 (70.54 vs.\ 64.15), Qwen-3VL (23.64 vs.\ 42.76), Qwen-2.5VL (19.88 vs.\ 16.74), and
DeepSeek (5.65 vs.\ 7.03). Counting appears to slightly increase difficulty for most of the
reasoning models, with Gemma-4 the notable exception. However, poor performance on
non-counting domains indicates that counting alone does not explain overall weaknesses.

Figure~\ref{fig:alt_skills} shows performance vs. size for these skills. Both counting and OCR exhibit sharp performance degradation with increasing size. In contrast, grid-based tasks show relatively stable performance, particularly for reasoning models (Gemma-4, GPT-5-mini, o4-mini, Gemini), with near-plateau trends. Notably, GPT-5-mini achieves 100\% accuracy in 3 out of 7 grid domains. This may reflect training data biases, as grid-like structures are common in tasks such as reinforcement learning with visual reasoning (RLVR).

\begin{figure}[!h]
\centering
\includegraphics[scale=0.5]{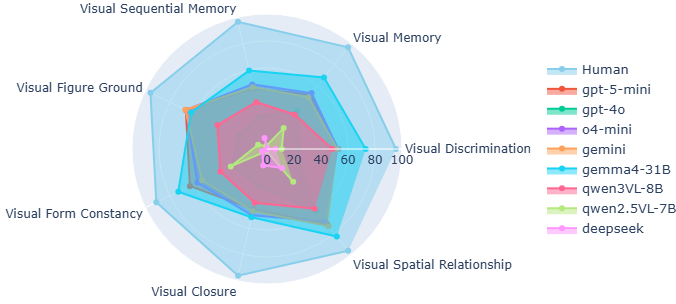}
\caption{MLLM average accuracy vs human accuracy across the 7 TVPS-4 skills.}
\label{fig:human_study}
\end{figure}

\subsection{Human Study}
\label{subsubsec:human_study}
% should answer Q5
We now answer Q5 through a human study where we recruit 16 college students, 6 female and 10 male. They were presented with randomly selected questions of any problem size from the 30 domains. Results of this study are presented in Figure~\ref{fig:human_study}. For fair comparison, the models are also evaluated on the exactly same samples, and exactly the same protocol as the humans. It can be seen that models lag severely as compared to humans. Even the best performance of the models exhibited by Gemini in Visual Spatial Relationship is around 73.08\% as compared to about 96.42\% accuracy shown by humans in the same skill. Overall (see Table \ref{table:human} in appendix) -- human performance is more than 26\% points higher than the closest MLLM. Interestingly, when testing on 'colours\_present', the domain on which all MLLMs have abysmal performance, we find the gap to be humongous -- humans achieve 90\% performance whereas the best MLLMs on the same questions achieve only 6.67\%.

\subsection{One-shot and Few-shot Results: }
We additionally perform one-shot experiments on proprietary models to assess how much exemplars can help MLLMs in understanding the task better, as well as to account for prompt sensitivity. Contrary to conventional wisdom, we find that one-shot prompting \emph{degrades} the overall performance,  with average of skills accuracy reducing to 51.25\% for GPT-5-mini, 17.62\% for GPT-4o, 52.26\% for o4-mini and 44.42\% for Gemini (refer to Table \ref{table:one_shot_results} in Appendix). Moreover, the skill-wise graphs also show degrading trends across sizes, similar to zero-shot (see Fig. \ref{fig:size_vs_acc_one_shot} in Appendix). Upon manual investigation of frequent errors made by models, we see that the models confuse the details of the in-context example with the problem to be solved. The models also commonly fail to perceive the visual details of the in-context examples and thus hallucinate wrong theories for solving the task. Further, a short few-shot experiment with 3 examples on one of the stronger proprietary models, Gemini showed that these mistakes recurred and the average accuracy on \dataset{} dropped to 21.5\%. Upon investigating model responses, we found that models frequently mixed details from the demonstration example with those of the target instance, particularly in tasks involving ordered outputs or spatial layouts. We believe this reveals an interesting limitation of current MLLMs in maintaining a clean separation between demonstration context and active image understanding.

% \SG{Upon investigating model responses, we found that models confuse the details of in-context examples with the problem to be solved.} This hints that MLLMs' weak performance is not likely because of task understanding, but due to inherent ability limitations.

\subsection{Additional Experimental Analysis}
\paragraph{Format Error Analysis: }
\label{sec:other}
Table \ref{table:answer_type_error} reports the accuracy of models in following answer formats - we see that proprietary models consistently follow answer formats while open source models are worse than them. The best performing is GPT-4o (with o4-mini close second) that sticks to the output formats with 0\% format errors.

\begin{table}[h]
\centering
\resizebox{0.8\columnwidth}{!}{%
    \begin{tabular}{|c|c|c|c|c|c|c|c|c|}
    \hline
    \bf{Answer Type} & \bf{GPT-5-mini} & \bf{GPT-4o} & \bf{o4-mini} & \bf{Gemini} & \bf{Gemma4} & \bf{Qwen3VL} & \bf{Qwen2.5VL} & \bf{DeepSeek}\\
    \hline
    single & 0.14 & 0.00 & 0.00 & 0.00 & 11.95 & 15.09 & 0.00 & 3.68\\
    fixed & 0.00 & 0.00 & 0.17 & 0.83 & 4.08 & 13.00 & 5.18 & 61.58\\
    list & 2.19 & 0.00 & 0.00 & 0.00 & 5.94 & 14.19 & 0.00 & 22.44\\
    set & 0.00 & 0.00 & 0.10 & 0.20 & 6.4 & 2.4 & 19.10 & 60.00\\
    %single & 100.00 & 100.00 & 100.00 & 100.00 & 96.32\\
    %fixed & 100.00 & 99.83 & 99.17 & 94.92 & 38.42\\
    %list & 100.00 & 100.00 & 100.00 & 100.00 & 77.56\\
    %set & 100.00 & 99.90 & 99.80 & 80.90 & 40.00\\
    \hline
    \end{tabular}%
}
\caption{Format Error (\%) Analysis of models}
\label{table:answer_type_error}
\end{table}

\paragraph{3D Percept-V: }
A potential criticism of Percept-V is that its images are auto-generated and may not occur naturally in model's training data. Will the models perform better if tested on more realistic images? To further investigate this concern, we developed 3D versions of a subset of Percept-V domains using Blender in the style of the CLEVR dataset~\citep{johnson2017clevr}. For six domains, we created 100 synthetic images spanning sizes 1–10 (10 images per size), where all 2D objects were converted into 3D objects with realistic lighting and shadows to mimic real-world photographs (Fig ~\ref{fig:3d_percept_v}). We present the results with GPT-5-mini and Qwen2.5-7B in Table~\ref{table:3D_results1} which shows that even for relatively small problem sizes-where performance on the corresponding 2D domains is comparatively high-performance on the 3D versions remains sub-par for most domains. Thus, as anticipated, simply introducing 3D rendering does not resolve the underlying perception challenges. Instead, these results point toward a deeper gap in the visual perceptual capabilities of current MLLMs.

\paragraph{Impact of Image Resolution: }

Percept-V images are already reasonably high resolution (670×650 pixels on average). To test whether resolution limits performance, we re-evaluated GPT-5-mini and Qwen on all images upscaled and downscaled by a factor of two. As reported in Table~\ref{table:resolution_results} upscaling does not provide much benefit: GPT-5-mini moves from 54.77 to 55.62 and Qwen from 14.64 to 15.01. This hints that the information required to solve these tasks was already present in the input and the failures are not caused by lack of clarity. Downscaling does hurt, as GPT-5-mini falls by 14 points and Qwen by 2.5 points, confirming that the tasks are resolution-sensitive in the expected direction and that the models are genuinely reading the images rather than guessing. Notably, the skills that suffer most under downscaling (Visual Memory and Visual Form Constancy) are those requiring many small objects to be individuated at once.

\begin{table}[h!]
\centering
\small
\begin{tabular}{|l|rrr|rrr|}
\hline
 & \multicolumn{3}{c|}{\bf GPT-5-mini} & \multicolumn{3}{c|}{\bf Qwen} \\
{\bf Skill} & {\bf Orig.} & {\bf $2\times$} & {\bf $0.5\times$} & {\bf Orig.} & {\bf $2\times$} & {\bf $0.5\times$} \\
\hline
Visual Discrimination      & 55.00 & 54.80 & 45.83 & 11.50 & 11.70 & 10.20 \\
Visual Memory              & 49.95 & 49.94 & 32.92 & 19.59 & 22.83 & 17.22 \\
Visual Sequential Memory   & 43.50 & 42.17 & 29.42 &  4.67 &  4.83 &  4.33 \\
Visual Figure Ground       & 64.75 & 63.12 & 52.00 &  7.58 &  4.88 &  4.62 \\
Visual Form Constancy      & 58.70 & 61.90 & 38.15 & 25.60 & 25.20 & 21.50 \\
Visual Closure             & 44.88 & 45.75 & 32.75 &  5.75 &  5.25 &  5.75 \\
Visual Spatial Relationship & 66.64 & 71.64 & 57.07 & 27.79 & 30.36 & 24.07 \\
\hline
Average of skills          & 54.77 & 55.62 & 41.16 & 14.64 & 15.01 & 12.53 \\
\hline
\end{tabular}
\caption{Accuracy (\%) at original resolution (670$\times$650 pixels on average),
upscaled $2\times$, and downscaled $0.5\times$.}
\label{table:resolution_results}
\end{table}

% \SG{
% The original Percept-V resolution was high with an average of 670x650 across all domains. However, we tried to analyse variation in performance with the change of resolution. So, we upscaled and downscaled all images by 2x and tested the performance of our best performing proprietary and open-source models, GPT-5-mini and Qwen on this dataset. Table \ref{table:resolution_results1} in section \ref{sec:appendix_resolution} lists the performance. From Table \ref{table:resolution_results} it is evident that upscaling or downscaling does not affect the performance of models too much with performance on upscaled images varying by only 1 percentage point for the average of skills from the performance on the original resolution dataset. Even though performance on upscaled images is slightly better than that on downscaled images, the difference is only about 14\% for GPT-5-mini and 3\% for Qwen. Hence, from this experiment it is clear that the poor performance is majorly because of gaps in perceptual understanding in present MLLMs.}

\paragraph{Impact of Test Phrases: }

We paraphrased the original prompts in two different styles - concise and verbose to understand the effects of prompt-styles on model performance. Table \ref{table:prompt_results} in Appendix shows the performance of GPT-5-mini and Qwen respectively on the different styles of prompts. From Table \ref{table:prompt_results}, it is clear that the style of text information does not affect the performance of models too much. In fact, the original Chain-of-Thought style prompting provides better results in many cases. Also, performance of concise prompts is mostly better than verbose prompts in case of both the models.

\section{Finetuning experiments} 
In this section we answer Q6 by exploring the performance of Qwen2.5-VL-7B when finetuned on Percept-V style data. We compare different configurations of LoRA \citep{DBLP:conf/iclr/HuSWALWWC22} and full-finetuning for both in-domain performance and out-of-domain generalization. To this end, we create a large version of Percept-V which we call Percept-V$_{\text{train}}$ for all finetuning based experiments, consisting of 60K samples with 2K samples for each domain. Evaluation is performed on Percept-V (in-domain) as well as CV-Bench~\citep{DBLP:conf/nips/TongBWWIAYYMWPF24} and Color-Bench~\citep{DBLP:conf/nips/LiangLFLNCBCLZ25} for out-of-domain assessment. We experiment with fine-tuning three different modules, the language encoder, vision encoder, and the projector (which aligns the two modalities). We perform these experiments with LoRA fine-tuning, and we also compare with fine-tuning the entire network. The results are presented in Table~\ref{table:fine-tuning}, with implementation details in ~\ref{sec:appendix_finetune}.

    \begin{table}[htbp]
    \centering
    \small 
    \resizebox{\linewidth}{!}{
    \begin{tabular}{ccc rrrrr}
    \toprule
    \textbf{LLM} & \begin{tabular}{@{}c@{}}\textbf{Vision} \\ \textbf{Encoder}\end{tabular} & \begin{tabular}{@{}c@{}}\textbf{MLP} \\ \textbf{Projector}\end{tabular} & \multicolumn{1}{c}{\begin{tabular}{@{}c@{}}\textbf{Percept-V} \\ \textbf{Overall}\end{tabular}} & \multicolumn{1}{c}{\begin{tabular}{@{}c@{}}\textbf{Percept-V} \\ \textbf{1-10} \\ \textbf{Problem Size}\end{tabular}} & \multicolumn{1}{c}{\begin{tabular}{@{}c@{}}\textbf{Percept-V} \\ \textbf{11-20} \\ \textbf{Problem Size}\end{tabular}} & \multicolumn{1}{c}{\begin{tabular}{@{}c@{}}\textbf{CV-Bench} \\ \textbf{Combined}\end{tabular}} & \multicolumn{1}{c}{\begin{tabular}{@{}c@{}}\textbf{Colorbench} \\ \textbf{Overall}\end{tabular}} \\
    \midrule
    \multicolumn{8}{c}{\textit{Baseline}}\\
    % \midrule
    \rowcolor{gray!20} $-$ & $-$ & $-$ & 18.10 & 29.81 & 06.35 & 73.92 & 46.28 \\
    \midrule
    \multicolumn{8}{c}{\textit{LoRA Fine-tuning}}\\
    \midrule
    $\square$ & $\square$ & $\checkmark$ & 49.15 & 65.05 & 33.28 & 74.02 & 47.43 \\
    $\square$ & $\checkmark$ & $\square$ & 68.67 & 80.76 & 56.60 & 68.49 & 46.49 \\
    $\checkmark$ & $\square$ & $\square$ & 75.18 & 84.33 & 66.05 & 76.70 & \textbf{47.54} \\
    $\square$ & $\checkmark$ & $\checkmark$ & 70.20 & 82.49 & 57.91 & 74.02 & 47.43 \\
    $\checkmark$ & $\square$ & $\checkmark$ & 74.63 & 83.80 & 65.49 & \textbf{78.06} & 42.72 \\
    $\checkmark$ & $\checkmark$ & $\square$ & 83.42 & 89.47 & \textbf{77.38} & 74.77 & 44.71 \\
    $\checkmark$ & $\checkmark$ & $\checkmark$ & 80.30 & 87.56 & 73.06 & 75.62 & 47.02 \\
    \midrule
    \multicolumn{8}{c}{\textit{Full Fine-tuning}}\\
    \midrule
    $\square$ & $\square$ & $\checkmark$ & 25.25 & 38.71 & 11.78& 74.46 &46.28\\
    $\square$ & $\checkmark$ & $\square$ & 75.80 & 87.5 & 64.09 & 72.53 &42.61\\
    $\checkmark$ & $\square$ & $\square$ & 82.60 & 92.60 & 72.57 & 58.91 &44.60\\
    $\square$ & $\checkmark$ & $\checkmark$ & 76.35 & 87.32 & 65.36 & 73.46 &44.08 \\
    $\checkmark$ & $\square$ & $\checkmark$ & 78.83 & 90.93 & 66.69 & 63.38 &44.71\\
    $\checkmark$ & $\checkmark$ & $\square$ & 84.18 & 93.55 & 74.79 & 69.54 &46.28\\
    $\checkmark$ & $\checkmark$ & $\checkmark$ & \textbf{84.30} & \textbf{93.33} & 75.27 & 77.56 & 45.97 \\
    \bottomrule
    % \midrule
    % \multicolumn{8}{c}{\textit{\textcolor{red}{Full Fine-tuning (5 epochs)}}}\\
    % \midrule
    % \rowcolor{red!20}$\square$ & $\square$ & $\checkmark$ & 25.10 & - & - & - &\\
    % \rowcolor{red!20}$\square$ & $\checkmark$ & $\square$ & 80.10 & - & - & - &\\
    % \rowcolor{red!20}$\checkmark$ & $\square$ & $\square$ & - & - & - & - &\\
    % \rowcolor{red!20}$\square$ & $\checkmark$ & $\checkmark$ & 80.83 & - & - & - &\\
    % \rowcolor{red!20}$\checkmark$ & $\square$ & $\checkmark$ & 88.50 & - & - & - &\\
    % \rowcolor{red!20}$\checkmark$ & $\checkmark$ & $\square$ & 89.25 & - & -& - &\\
    % \rowcolor{red!20}$\checkmark$ & $\checkmark$ & $\checkmark$ & \textbf{90.77} & - & - & - &\\
    % \bottomrule
    \end{tabular}}
    \caption{\textbf{Legend:} \textbf{L}: LLM, \textbf{V}: Vision Encoder, \textbf{P}: MLP Projector between LLM and Vision Encoder. Ordering in LoRA style Fine-tuned models (overall accuracy on \dataset{}): (L+V) $>$ (L+V+P) $>$ (L) $>$ (L+P) $>$ (V+P) $>$ (V) $>$ (P)}
    % \vspace{1.5ex}
    % \raggedright
    % \footnotesize
    \label{table:fine-tuning}
    \end{table}

It is evident from Table~\ref{table:fine-tuning} that both LoRA and full fine-tuning yields significant gains (over 60pts) over the pre-trained model.
Interestingly, LoRA achieves performance only marginally worse than full-fine tuning. We see most of the gains coming from training the language encoder (LLM), followed by the vision encoder, which means that training even one of the modalities has a significant boost on performance. Fine-tuning both the modalities together results in further gains. Surprisingly, adding the projector to the set of modules being fine-tuned, does not benefit (last but one row in LoRA experiments), and we believe this may be due to the need to have a more careful curriculum during fine-tuning. This may further imply that models struggle to query, align and transform visual information and motivates stronger interfaces between vision and language modalities.

Table~\ref{table:fine-tuning} also compares the performance across the test data divided across problems of sizes 1-10, and 11-20. Naturally, we observe better performance on smaller problems which is expected. Interestingly, the performance gap decreases from 23 points (first row), to 18 points (last row) for the two subsets of the data, after full-fine tuning. This means that fine-tuning is able to address some aspects of the hardness of the problem, resulting from increasing size. For most part, the fine-tuning benefits do not significantly carry over to other similar datasets, which are on real images but require basic perceptual abilities similar to ours,  with only a marginal increase in one of the datasets (though there is no catastrophic forgetting either). This likely highlights the fact that fine-tuned model is primarily able to learn the patterns on the given task, and is not fully able to generalize the learned ability to other tasks requiring similar skills.

\section{Conclusion}
%Our paper takes an important step forward in assessing the perceptual abilities of multimodal LLMs. 
We present \dataset{}, an extensive dataset of 6000 automatically generated images from 30 domains where each domain is mapped with a subset of perceptual skills from the TVPS-4 Cognitive Science framework.  
%To isolate perceptual ability, questions in our dataset makes use of very few basic reasoning skills (such as counting), and similarly very basic knowledge (of colors and basic shapes). 
A variety of open and closed source MLLMs perform rather weakly on almost all domains, with best average performance of any model being 66.9\%, demonstrating significant gap in their ability to solve basic perceptual tasks. Our human study experiments further show that a significant gap exists between human and LLM performance on very simple perception tasks in our dataset. Interestingly, fine-tuning on our dataset improves the performance, though does not help with generalization on other similar datasets. Ablation studies suggest that visual representation and multimodal models both require fine tuning for best performance,
pointing to the limitations of the underlying representation and alignment. Two failure mode recur throughout our experiments on pre-trained models: accuracy falls steadily as problem size grows and in-context examples hurt rather than help. Both point to issues with visual capacity and the handling of multi-image context as the limitations that must be addressed to improve basic perception. Our data and code are available at \href{https://dair-iitd.github.io/Percept-V/}{https://dair-iitd.github.io/Percept-V/}.

%the gap between the best MLLM and humans is quite high suggesting that contrary to LLMs' high performance against humans in existing tasks (e.g., LLMs winning Math olympiad), MLLMs are rather behind, especially on basic perception skills. Even though one-shot and few-shot prompting degrade the model accuracy, finetuning Qwen-2.5-VL on our dataset significantly improves its performance. We will release the full \dataset{} dataset upon publication, and hope that it will become a de-facto standard for benchmarking perceptual abilities of any new and existing MLLMs.

\section*{Ethics Statement}
All human participants were shown the instructions required to be followed during the experiment, and they volunteered to participate in the study. Additionally, all humans were asked for consent and financially compensated for their contribution -- details can be found in the \ref{section:participant_instructions} of Appendix \ref{sec:appendix}. Our work only tries to expose the perceptual limitations of SoTA MLLMs, so human study did not elicit any personally identifying information -- the participants were simply answering simple perception questions. 

\section*{Acknowledgements}
Mausam is supported by grants from Google, Microsoft, Wipro, Verisk, Tower Research and Nick
McKeown Chair Professorship. Parag Singla is supported by
Shanthi and K Ananth Krishnan Young Faculty
Chair Professorship. Parag and Mausam are supported by the IBM AI Horizon Networks (AIHN) grant. We thank IIT Delhi HPC compute facility, Microsoft AFMR program and Google Gemini
Credits grant for accesss to compute and LLM API calls, Yardi School of Artificial Intelligence for travel grant. Any opinions, findings, conclusions or recommendations expressed here are those of the authors and do not necessarily reflect the views
or official policies of the funding agencies.

\bibliographystyle{colm2026_conference}
\bibliography{colm2026_conference}

\newpage

\appendix
\section{Appendix}
\label{sec:appendix}

\subsection{Domain examples of MLLMs} 

\subsubsection{Change Color} 
\textbf{Input Prompt:} \\
-These are two identical images with circles of different colours.
- The colours of the circles at the same position in the two images may be different.
- Count the number of differently coloured circles between the two images.
- The output must be given in a single line in the form of COUNT:x

\begin{figure}[!h]
    \centering
    % --- LEFT SIDE: TABLE ---
    \begin{minipage}{0.45\columnwidth}
        \centering
        \frame{\includegraphics[scale=0.12]{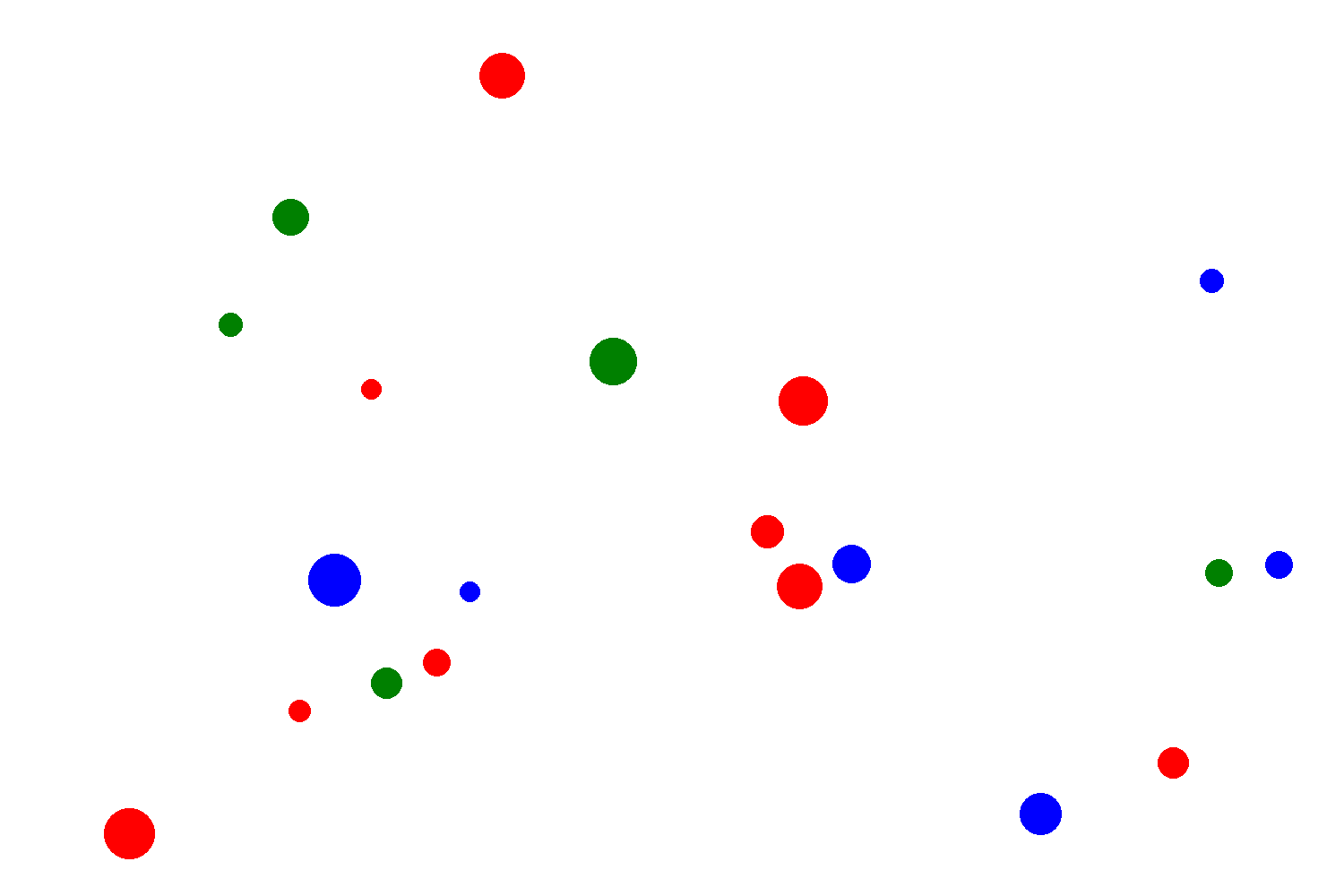}}
        \caption{Change Color First Image}
    \end{minipage}
    \hfill % Adds space between minipages
    % --- RIGHT SIDE: FIGURE ---
    \begin{minipage}{0.45\columnwidth}
        \centering
        \frame{\includegraphics[scale=0.12]{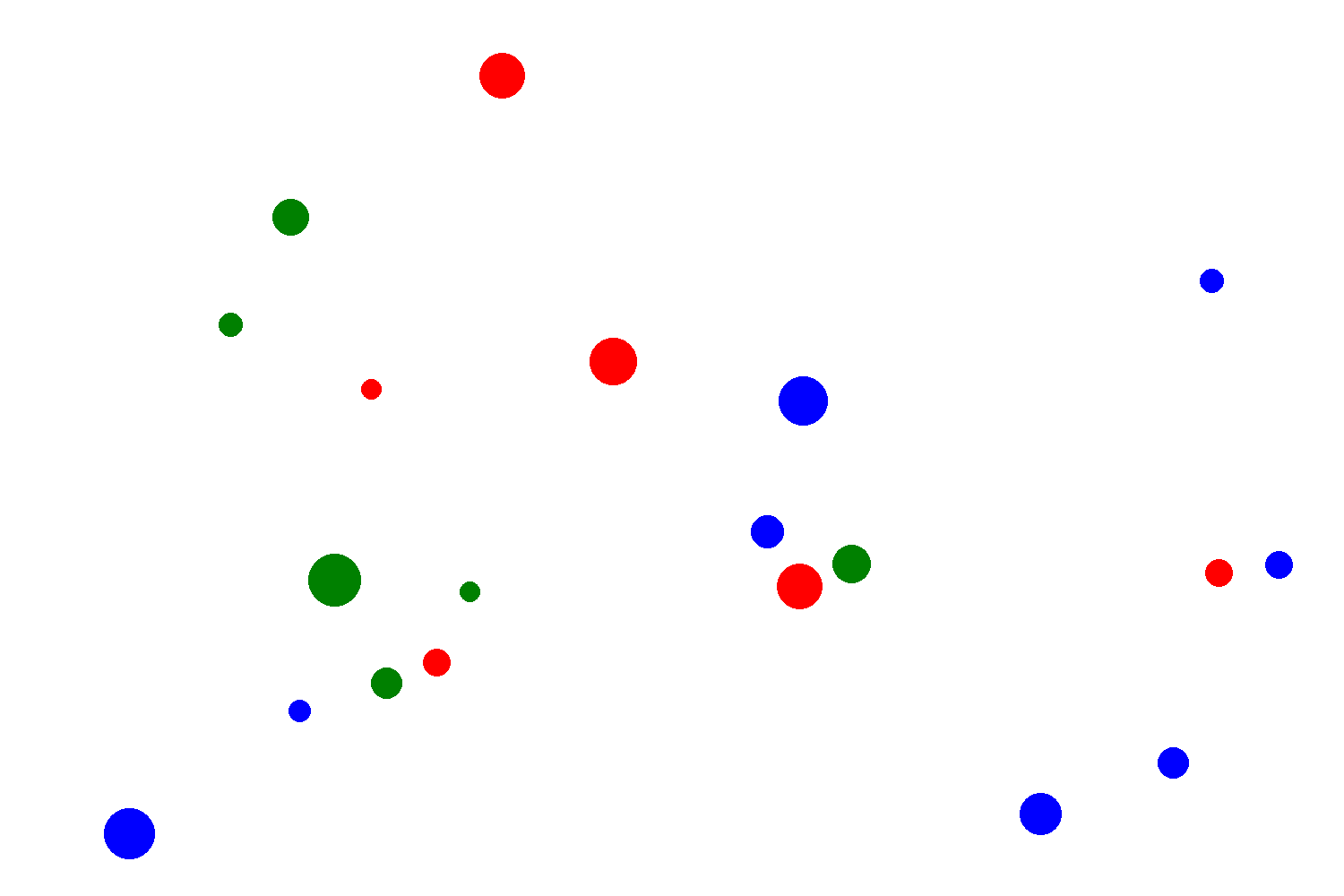}}
        \caption{Change Color Second Image}
    \end{minipage}
\end{figure}

\subsubsection{Circle Boxes} 
\textbf{Input Prompt:} \\
- This is an image with some red circles divided and a blue partition that divides the image into two sides.
- Some circles are on the left side of the partition whereas some are on the right. Note that, a side of the image may be completely empty as well.
- Determine the number of circles to be moved from the side with more number of circles to the side with less number of circles so that each side has equal number of circles.
- In case of both the sides having equal number of circles, return zero.
- The output must be given in a single line in the form of COUNT:x where x is the number of circles to be transferred

\subsubsection{Circle Location} 
\textbf{Input Prompt:} \\
- This is an image with some black circles drawn in the four quadrants
- The quadrants are numbered in the default way i.e. quadrant 1 is the rightmost and topmost quadrant, quadrant 2 is the leftmost and topmost, quadrant 3 is the leftmost and lowest and quadrant 4 is the rightmost and lowest.
- Determine the quadrant with the most number of black circles and the count of black circles in that quadrant.
- In case of a tie, return the lesser quadrant number.
- The output must be given in a single line in the form of QUADRANT:x COUNT:y where x is the quadrant number and y is the count of circles.

\begin{figure}[!h]
    \centering
    % --- LEFT SIDE: TABLE ---
    \begin{minipage}{0.45\columnwidth}
        \centering
        \frame{\includegraphics[scale=0.5]{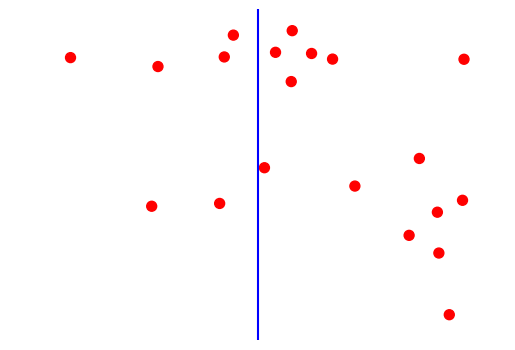}}
        \caption{Circle Boxes}
    \end{minipage}
    \hfill % Adds space between minipages
    % --- RIGHT SIDE: FIGURE ---
    \begin{minipage}{0.45\columnwidth}
        \centering
        \frame{\includegraphics[scale=0.5]{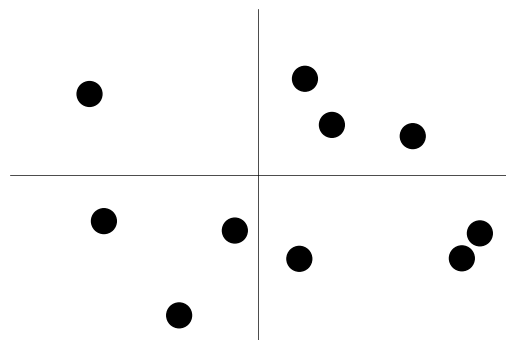}}
        \caption{Circle Location}
    \end{minipage}
\end{figure}

\subsubsection{Circle Right Triangle} 
\textbf{Input Prompt:} \\
- This is a grid which contains some circles and triangles
- Check if any triangle cell has a circle cell to its immediate right.
- The last line of output should be "NO" if no such cell exists and "YES" otherwise.

\begin{figure}[!h]
    \centering
    % --- LEFT SIDE: TABLE ---
    \begin{minipage}{0.45\columnwidth}
        \centering
        \includegraphics[scale=0.4]{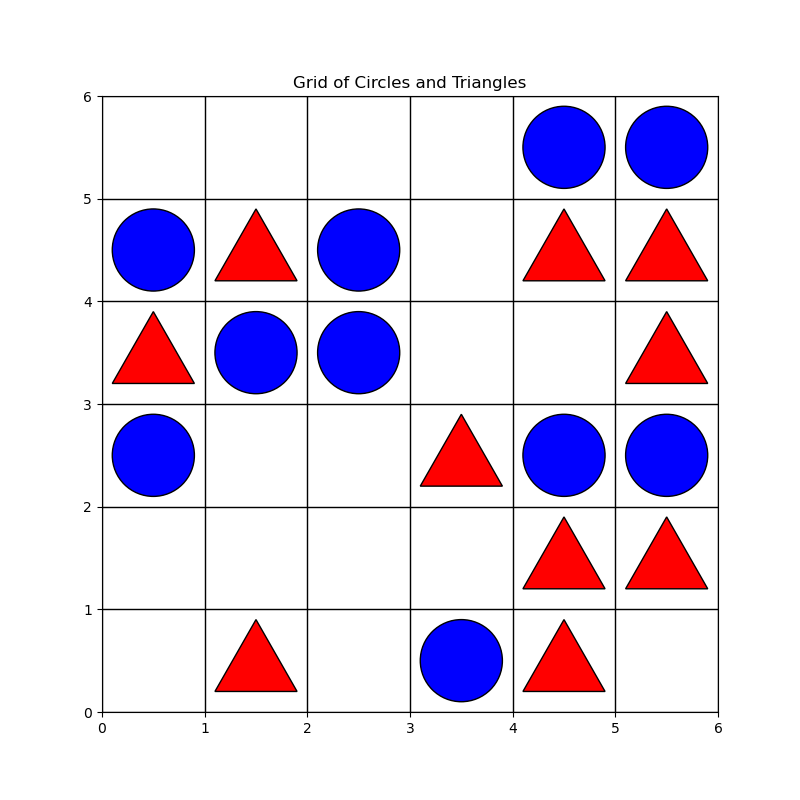}
        \caption{Circle Right Triangle}
    \end{minipage}
    \hfill % Adds space between minipages
    % --- RIGHT SIDE: FIGURE ---
    \begin{minipage}{0.45\columnwidth}
        \centering
        \frame{\includegraphics[scale=0.35]{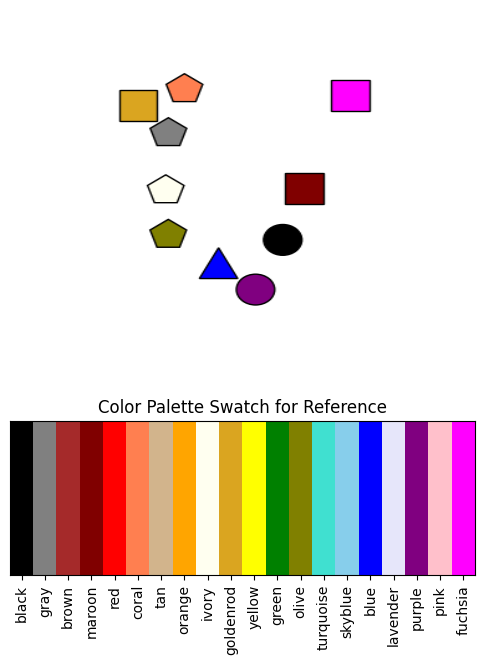}}
        \caption{Colours Present}
    \end{minipage}
\end{figure}

\subsubsection{Colours Present} 
\textbf{Input Prompt:} \\
- This image contains some shapes in different colours. 
- The colours belong to the following list - [black, gray, brown, maroon, red, coral, tan, orange, ivory, goldenrod, yellow, green, olive, turquoise, skyblue, blue, lavender, purple, pink, fuchsia]
- No colour is repeated
- Produce a list of yes or no; write yes in the list if the colour at the corresponding index of the above list is present and no if it is absent.
- The output should be written in a single line as a comma-separated list of yes or no, written in order of the given list of colours.
- For example, the output should be of the form ANSWER: yes , no, no, yes and so on.

\subsubsection{Comparing Size} 
\textbf{Input Prompt:} \\
- This is an image with several rows of circles where each row is above a horizontal line.
- The image may also contain a single row with a blue and a green circle.
- Each row contains two colored circles (one blue and one green) of different sizes.
- Analyse which circle is bigger between the two in each row.
- The last line of the output should be of form "ANSWER:" followed by the space separated color answers for all the rows in the image.
- For example ANSWER: Blue Green Blue

\subsubsection{Count Colored Circles} 
\textbf{Input Prompt:} \\
- This is an image containing some colored circles
- Count the total number of red circles in the image
- Last line of the output must be of the form COUNT:x

\begin{figure}[!h]
    \centering
    % --- LEFT SIDE: TABLE ---
    \begin{minipage}{0.45\columnwidth}
        \centering
        \frame{\includegraphics[scale=0.7]{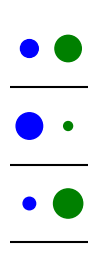}}
        \caption{Comparing Size}
    \end{minipage}
    \hfill % Adds space between minipages
    % --- RIGHT SIDE: FIGURE ---
    \begin{minipage}{0.45\columnwidth}
        \centering
        \frame{\includegraphics[scale=0.4]{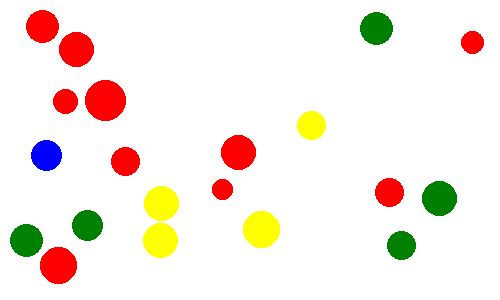}}
        \caption{Count Coloured Circles}
    \end{minipage}
\end{figure}

\subsubsection{Counting Circles} 
\textbf{Input Prompt:} \\
- This is an image containing some black circles
- Count the total number of circles in the image
- The last line of the output should be of form COUNT:x.

\subsubsection{Counting Locations} 
\textbf{Input Prompt:} \\
- There will be an image containing a rectangle and some circles above and below it
- Count the number of circles above the rectangle and below the rectangle
- The output must be of form in a single line ABOVE:x BELOW:y

\begin{figure}[!h]
    \centering
    % --- LEFT SIDE: TABLE ---
    \begin{minipage}{0.45\columnwidth}
        \centering
        \frame{\includegraphics[scale=0.4]{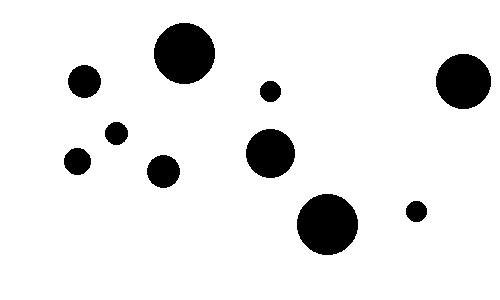}}
        \caption{Counting Circles}
    \end{minipage}
    \hfill % Adds space between minipages
    % --- RIGHT SIDE: FIGURE ---
    \begin{minipage}{0.45\columnwidth}
        \centering
        \frame{\includegraphics[scale=0.4]{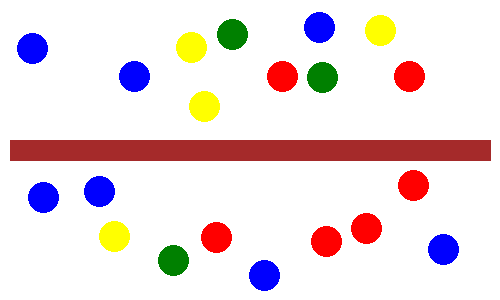}}
        \caption{Count Counting Locations}
    \end{minipage}
\end{figure}

\subsubsection{Counting Shapes} 
\textbf{Input Prompt:} \\
- This is an image containing shapes such as circles, triangles and squares
- Count the number of circles, triangles and sqaures in the image
- The last line of the output must be of the form CIRCLES:x TRIANGLES:y SQUARES:z

\subsubsection{Cross and Knots} 
\textbf{Input Prompt:} \\
- This is a grid containing some black X.
- Each cell is given a coordinate (x,y) which is (row,column) with 0 based indexing.
- An adjacent cell to a cell A is any cell sharing an edge with A
- A safe cell is a cell that is not adjacent to any black X.
- Output the coordinates of any one safe cell as labelled in the grid.
- If no such cell exists output None.
- The last line of the output should just contain the cell coordinates (x,y) or None and nothing else.

\begin{figure}[!h]
    \centering
    % --- LEFT SIDE: TABLE ---
    \begin{minipage}{0.45\columnwidth}
        \centering
        \frame{\includegraphics[scale=0.35]{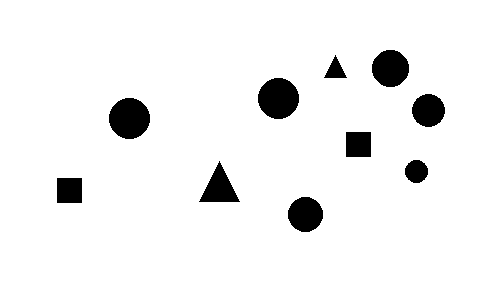}}
        \caption{Counting Shapes}
    \end{minipage}
    \hfill % Adds space between minipages
    % --- RIGHT SIDE: FIGURE ---
    \begin{minipage}{0.45\columnwidth}
        \centering
        \includegraphics[scale=0.5]{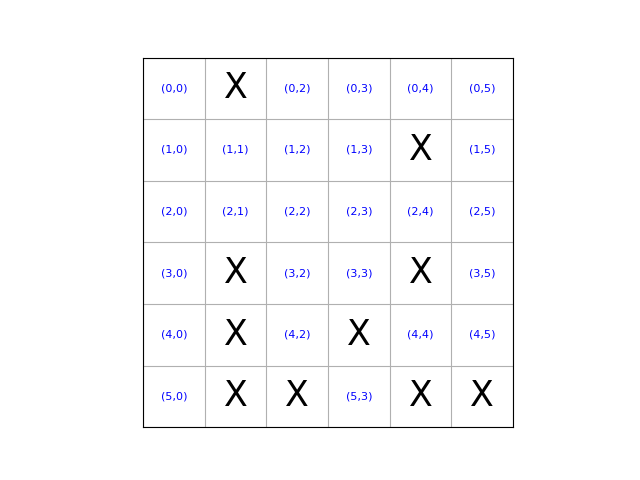}
        \caption{Cross and Knots}
    \end{minipage}
\end{figure}

\subsubsection{Graph Counting} 
\textbf{Input Prompt:} \\
- This is a graph where the blue colored circles are nodes and the lines joining them are edges.
- The graph may contain a single blue node with no edges.
- The graph may be disconnected.
- Count the number of nodes and edges in the graph.
- Last line of the output must be of the form NODES:x EDGES:y.

\subsubsection{Grid Path} 
\textbf{Input Prompt:} \\
- This is a grid which contains some circles, triangles and squares.
- Travel from cell labelled S to cell labelled E.
- Follow the path marked by the black line and arrows.
- Write down the sequence of shapes seen while moving along this path.
- The shapes present in the S and E cells should also be included in the sequence.
- The last line of the output should be a comma-separated list of shapes without colors, written in order of visiting of form SHAPES : Shape 1 , Shape 2 and so 
on.

\begin{figure}[!h]
    \centering
    % --- LEFT SIDE: TABLE ---
    \begin{minipage}{0.45\columnwidth}
        \centering
        \frame{\includegraphics[scale=0.22]{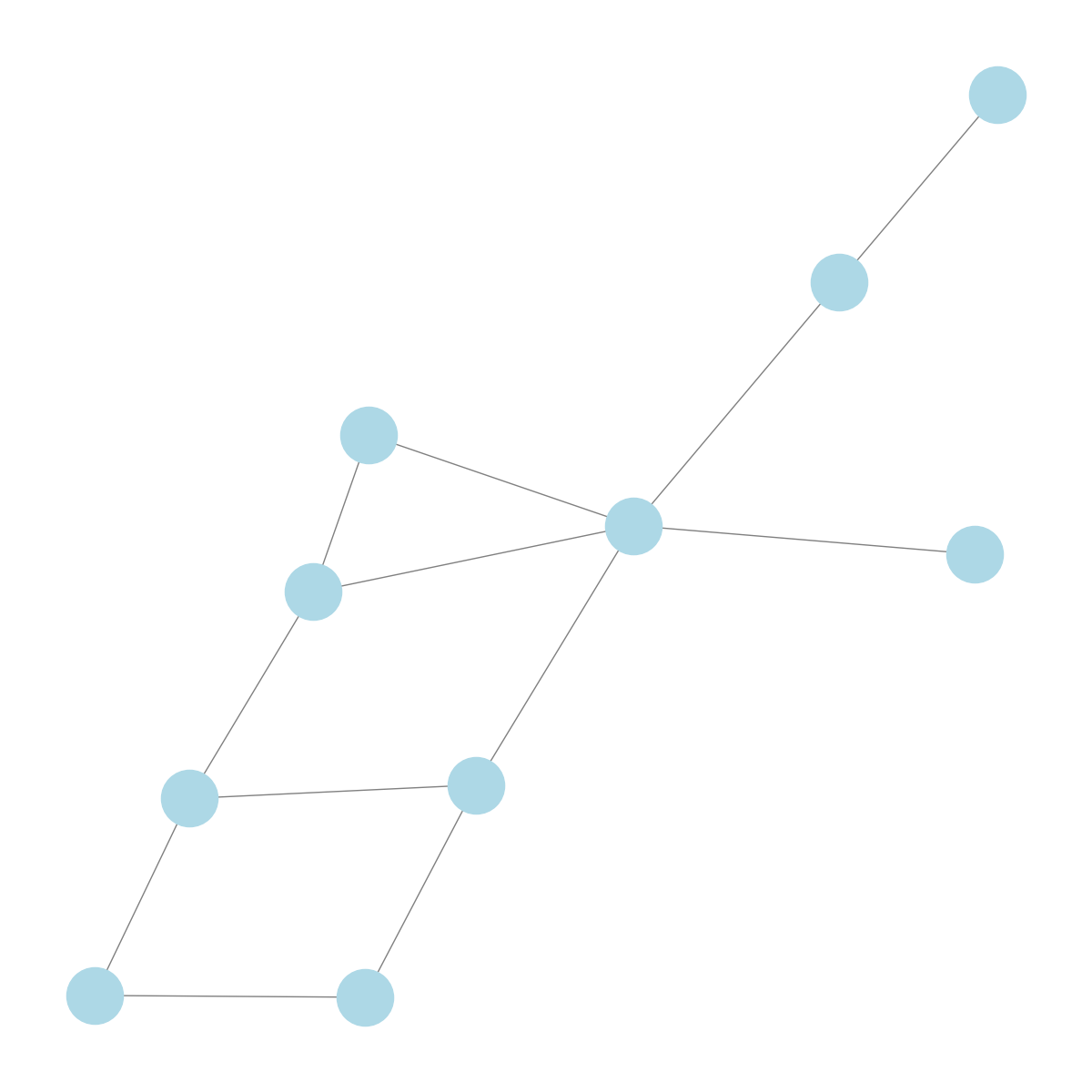}}
        \caption{Graph Counting}
    \end{minipage}
    \hfill % Adds space between minipages
    % --- RIGHT SIDE: FIGURE ---
    \begin{minipage}{0.45\columnwidth}
        \centering
        \includegraphics[scale=0.25]{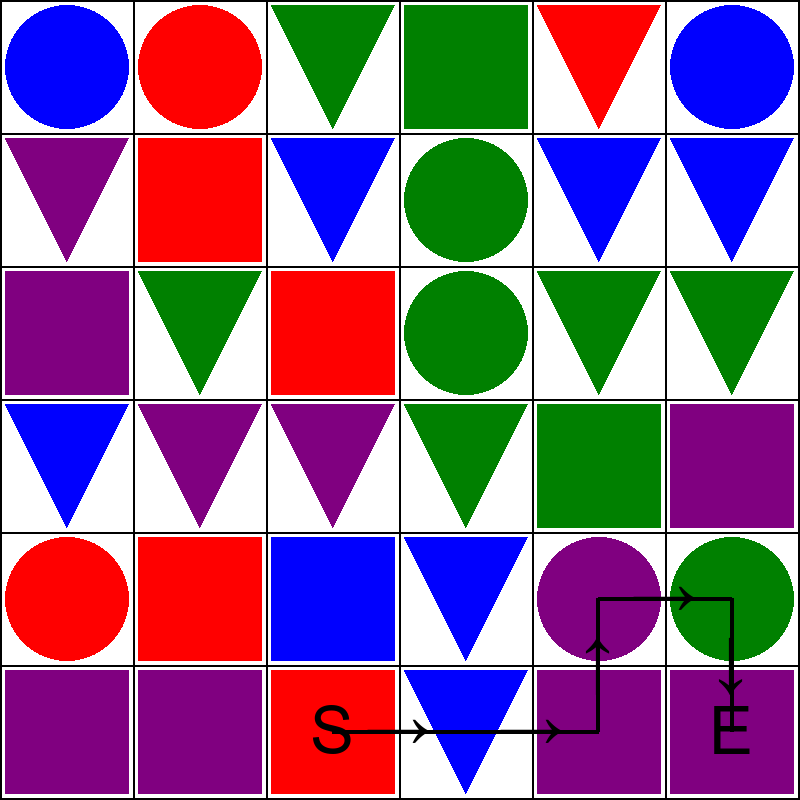}
        \caption{Grid Path}
    \end{minipage}
\end{figure}

\subsubsection{Identifying Shapes} 
\textbf{Input Prompt:} \\
- This is an image with several rows of different shapes where each row is above a horizontal line.
- Each row contains a coloured shape.
- The image may also contain a single shape.

- Identify the shape present in each row.
- In case there is a single shape, treat it as a single row.
- The ouput should have only a single line containing the answer.
- The output format should be "ANSWER:" followed by the space seperated shape names.
- For example, ANSWER: Circle Triangle Circle

\subsubsection{Inside Circles} 
\textbf{Input Prompt:} \\
- This is an image containing some black circles and a single red dot
- Determine if the red dot is contained inside any circle in the image
- Note that a circle containing the red dot means the centre of the dot is present inside the circumference of the circle
- The last line of the output should be of form ANSWER: x where x is Yes if the dot is contained in any circle and No otherwise.

\begin{figure}[!h]
    \centering
    % --- LEFT SIDE: TABLE ---
    \begin{minipage}{0.45\columnwidth}
        \centering
        \frame{\includegraphics[scale=0.75]{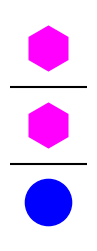}}
        \caption{Identifying Shapes}
    \end{minipage}
    \hfill % Adds space between minipages
    % --- RIGHT SIDE: FIGURE ---
    \begin{minipage}{0.45\columnwidth}
        \centering
        \frame{\includegraphics[scale=0.5]{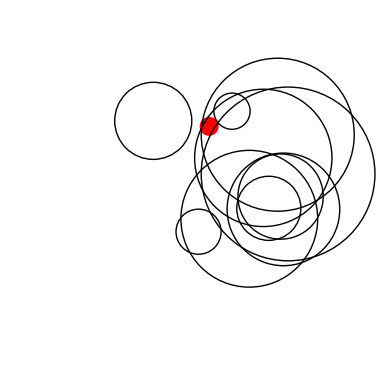}}
        \caption{Inside Circles}
    \end{minipage}
\end{figure}

\subsubsection{Layered Colors} 
\textbf{Input Prompt:} \\
- This image contains some concentric circles in different colours. 
- The image may also contain a single circle
- Write down the sequence of colours seen from inside to outside.
- In case of a single circle in the image, write down the colour of that circle.
- The colours belong to the following list - [black, gray, brown, maroon, red, coral, tan, orange, ivory, goldenrod, yellow, green, olive, turquoise, skyblue, blue, lavender, purple, pink, fuchsia]
- No colour is repeated
- The output should be written in a single line as a comma-separated list of colours, written in order from the innermost colour to the outermost colour        
- For example, the output should be of the form COLOURS: Colour 1 , Colour 2 and so on.

\subsubsection{Layered Shapes} 
\textbf{Input Prompt:} \\
- This image contains some shapes that are layered on top of each other in a white background.
- The shapes colud be of the following types - [circle, diamond, hexagon, octagon]
- The image could also contain a single shape with no other shapes layered on top of it.
- Write down the sequence of shapes from the inside layer to the outside layer.
- If there is a single shape then identify it.
- The last line of the output should be a comma-separated list of shapes without colors, written in order from top to bottom of form SHAPES : Shape 1 , Shape 2 
and so on.
- If there is a single shape then write SHAPES : shape-name where shape-name is the name of the shape seen

\begin{figure}[!h]
    \centering
    % --- LEFT SIDE: TABLE ---
    \begin{minipage}{0.45\columnwidth}
        \centering
        \frame{\includegraphics[scale=0.27]{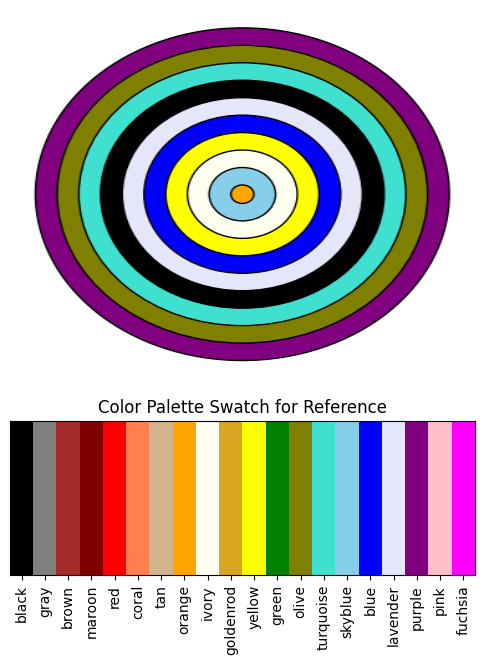}}
        \caption{Layered Colours}
    \end{minipage}
    \hfill % Adds space between minipages
    % --- RIGHT SIDE: FIGURE ---
    \begin{minipage}{0.45\columnwidth}
        \centering
        \frame{\includegraphics[scale=0.1]{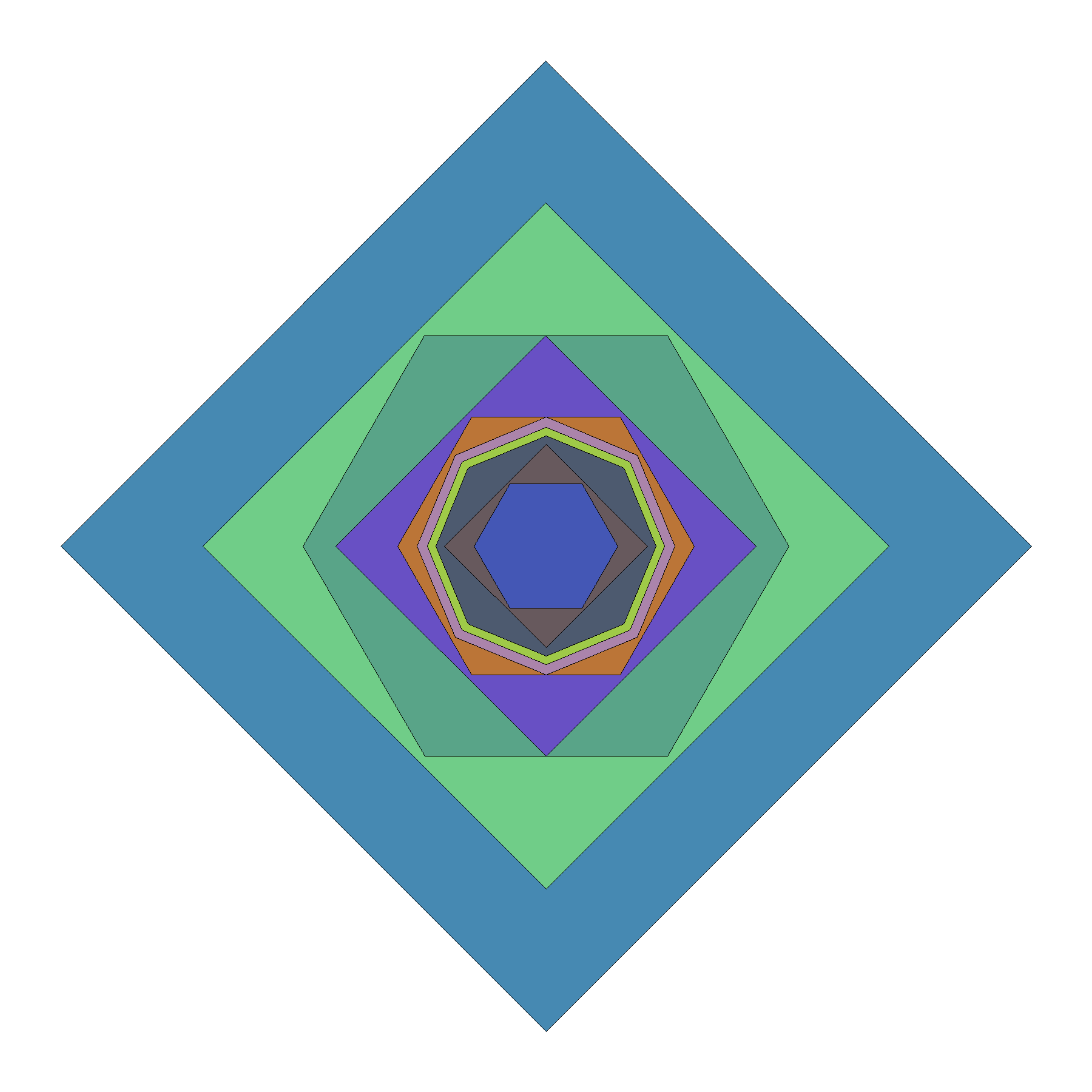}}
        \caption{Layered Shapes}
    \end{minipage}
\end{figure}

\subsubsection{List Colors} 
\textbf{Input Prompt:} \\
- This is an image with some multi-coloured shapes drawn on top of it
- The colours belong to the following list - [black, gray, brown, maroon, red, coral, tan, orange, ivory, goldenrod, yellow, green, olive, turquoise, skyblue, blue, lavender, purple, pink, fuchsia]
- No colour is repeated
- Identify the colour of the shapes drawn over the background image
- Form a list of all the colours of the shapes
- The output must be given in a single line in the form of a list of all the colours found

\subsubsection{List Shapes} 
\textbf{Input Prompt:} \\
- This is an image with some multi-coloured shapes drawn on top of it
- The shapes belong to the following list - [circle, triangle, square, pentagon]
- Identify the shapes drawn over the background image
- Form a set of all the unique shapes seen in the image
- The output must be given in a single line in the form of a set of unique shapes found

\begin{figure}[!h]
    \centering
    % --- LEFT SIDE: TABLE ---
    \begin{minipage}{0.45\columnwidth}
        \centering
        \frame{\includegraphics[scale=0.3]{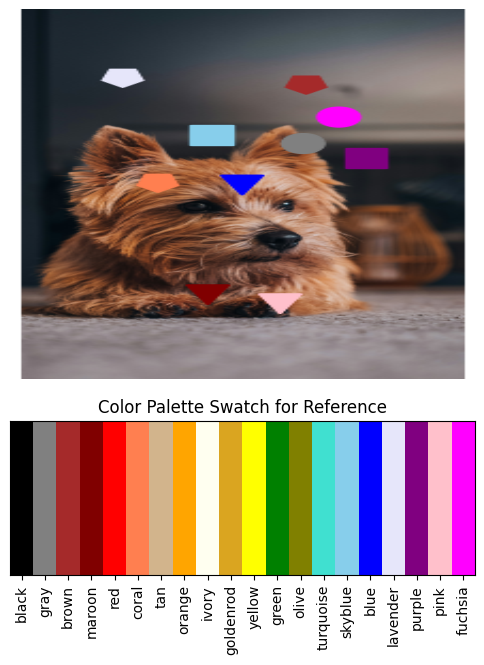}}
        \caption{List Colours}
    \end{minipage}
    \hfill % Adds space between minipages
    % --- RIGHT SIDE: FIGURE ---
    \begin{minipage}{0.45\columnwidth}
        \centering
        \frame{\includegraphics[scale=0.75]{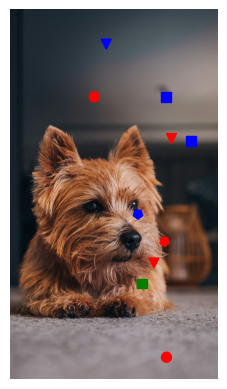}}
        \caption{List Shapes}
    \end{minipage}
\end{figure}

\subsubsection{Locate Circles Color} 
\textbf{Input Prompt:} \\
- This is a grid which contains some coloured circles.
- The rows and columns are numbered in the grid.
- The coordinate of a cell is given by the row number followed by the column number.
- Write down all the coordinates of the cells which contain a green circle.
- The last line of the output should be a space-separated list of the coordinates of the cells which contain a green circle.
- The coordinates should be in the format (row,column) with the comma and bracket

\subsubsection{Locate Circles Shape} 
\textbf{Input Prompt:} \\
- This is a grid which contains some coloured circles, rectangles and triangles.
- The rows and columns are numbered in the grid.
- The coordinate of a cell is given by the row number followed by the column number.
- Write down all the coordinates of the cells which contain a green circle.
- The last line of the output should be a space-separated list of the coordinates of the cells which contain a green circle.
- The coordinates should be in the format (row,column) with the comma and bracket.

\begin{figure}[!h]
    \centering
    % --- LEFT SIDE: TABLE ---
    \begin{minipage}{0.45\columnwidth}
        \centering
        \includegraphics[scale=0.3]{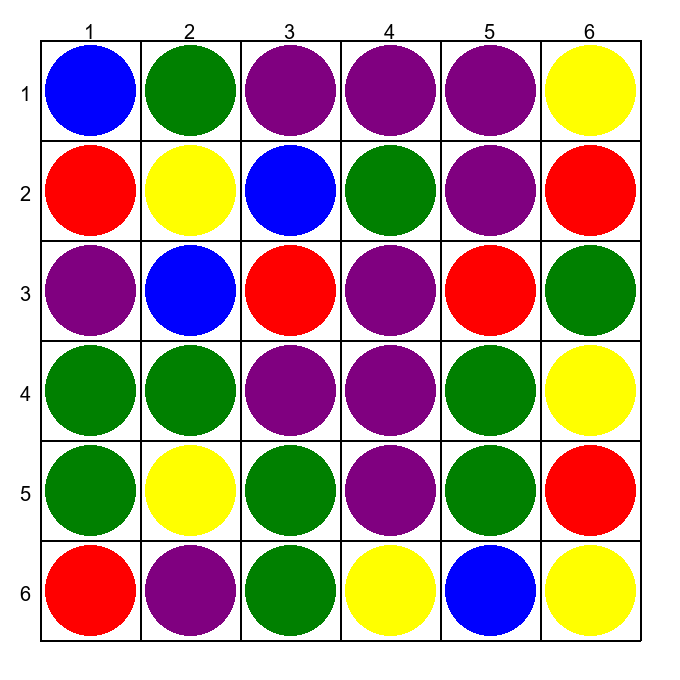}
        \caption{Locate Circles Colour}
    \end{minipage}
    \hfill % Adds space between minipages
    % --- RIGHT SIDE: FIGURE ---
    \begin{minipage}{0.45\columnwidth}
        \centering
        \includegraphics[scale=0.3]{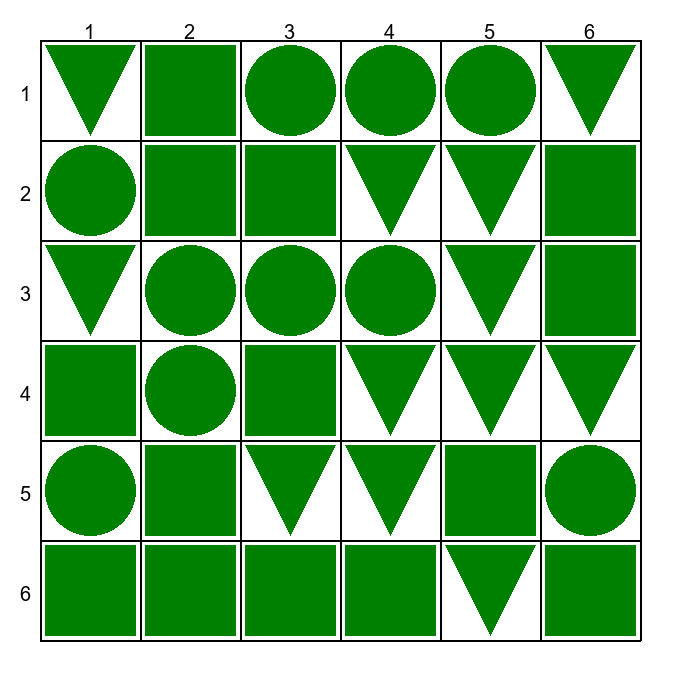}
        \caption{Locate Circles Shapes}
    \end{minipage}
\end{figure}

\subsubsection{Match Outline} 
\textbf{Input Prompt:} \\
- These are two images of some sequence of shapes separated by black horizontal lines
- The two images may also contain a single shape each
- The first image consists of rows of shapes and the second image consists of rows of outlines of shapes
- Determine the number of shapes in the first image that match the outlines in the corresponding row in the second image.
- In case each of the two images contain only a single shape, treat the single shape in each image as a row.
- The output must be given in a single line in the form of ANSWER:x where x is the number of matching shapes

\subsubsection{Match Shadow} 
\textbf{Input Prompt:} \\
- These are two images of some sequence of shapes separated by black horizontal lines
- The two images may also contain a single shape each
- The first image consists of rows of shapes and the second image consists of rows of shadows of those shapes.
- Determine the number of shapes in the first image that match the shadows in the corresponding row in the second image.
- In case each of the two images contain only a single shape, treat the single shape in each image as a row.
- The output must be given in a single line in the form of ANSWER:x where x is the number of matching shadows

\begin{figure}[!h]
    \centering
    % --- LEFT SIDE: TABLE ---
    \begin{minipage}{0.23\columnwidth}
        \centering
        \frame{\includegraphics[scale=0.7]{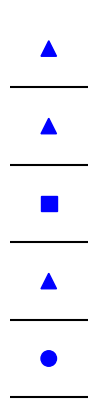}}
        \caption{Match Outline 1st Image}
    \end{minipage}
    \hfill % Adds space between minipages
    % --- RIGHT SIDE: FIGURE ---
    \begin{minipage}{0.23\columnwidth}
        \centering
        \frame{\includegraphics[scale=0.7]{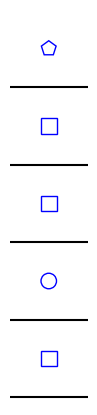}}
        \caption{Match Outline 2nd Image}
    \end{minipage}
    \hfill
     % --- LEFT SIDE: TABLE ---
    \begin{minipage}{0.23\columnwidth}
        \centering
        \frame{\includegraphics[scale=0.7]{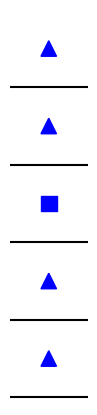}}
        \caption{Match Shadow 1st Image}
    \end{minipage}
    \hfill % Adds space between minipages
    % --- RIGHT SIDE: FIGURE ---
    \begin{minipage}{0.23\columnwidth}
        \centering
        \frame{\includegraphics[scale=0.7]{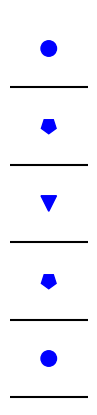}}
        \caption{Match Shadow 2nd Image}
    \end{minipage}
\end{figure}

\subsubsection{Maze Solving} 
\textbf{Input Prompt:} \\
- This is a maze.
- You start from the cell labelled S.
- Your goal is to reach cell labelled E through the shortest path possible.
- You can move up, down, left or right.
- You cannot move through the black walls.
- You can only move through white cells.
- Give the sequence of cells taken to go from S to E.
- Last line of output should be a comma separated line with cell numbers starting with S and ending at E and nothing else.

\begin{figure}[!h]
    \centering
    % --- LEFT SIDE: TABLE ---
    \begin{minipage}{0.45\columnwidth}
        \centering
        \includegraphics[scale=0.45]{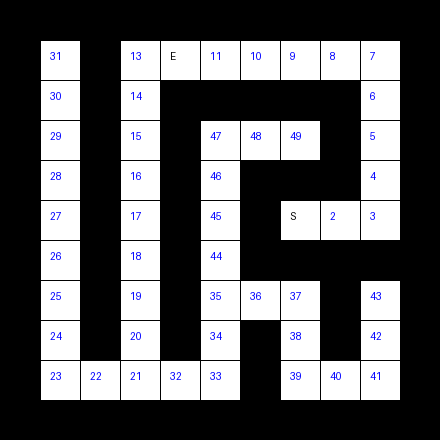}
        \caption{Maze Solving}
    \end{minipage}
    \hfill % Adds space between minipages
    % --- RIGHT SIDE: FIGURE ---
    \begin{minipage}{0.45\columnwidth}
        \centering
        \frame{\includegraphics[scale=0.5]{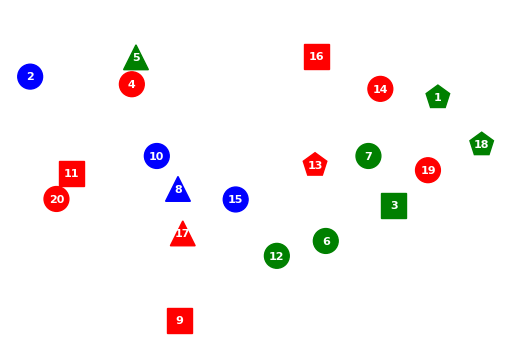}}
        \caption{Numbered Shapes}
    \end{minipage}
\end{figure}

\subsubsection{Mirror Image} 
\textbf{Input Prompt:} \\
- These are two images containing different shapes in different colours
- The image consists of a vertical blue line at the centre.
- Some objects of the first image are mirrored along the given vertical line in the second image.
- Other objects in the second image maintain the same postions as in the first image.
- Determine the number of objects that are mirrored in the second image.
- The output must be given in a single line in the form of COUNT: x, where x is the number of being mirrored
\begin{figure}[!h]
    \centering
    % --- LEFT SIDE: TABLE ---
    \begin{minipage}{0.45\columnwidth}
        \centering
        \frame{\includegraphics[scale=0.55]{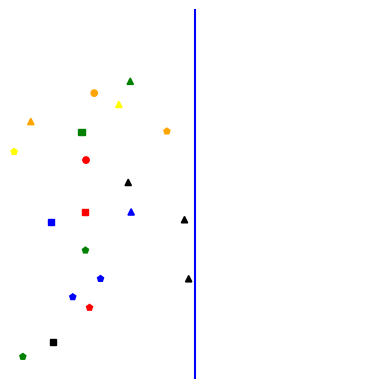}}
        \caption{Mirror Image First Image}
    \end{minipage}
    \hfill % Adds space between minipages
    % --- RIGHT SIDE: FIGURE ---
    \begin{minipage}{0.45\columnwidth}
        \centering
        \frame{\includegraphics[scale=0.55]{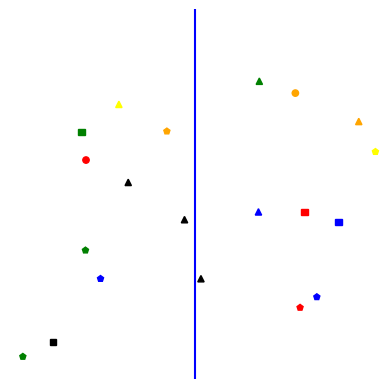}}
        \caption{Mirror Image Second Image}
    \end{minipage}
\end{figure}

\subsubsection{Numbered Images} 
\textbf{Input Prompt:} \\
- This is an image with some shapes drawn and numbers written on top of the shapes
- Determine the list of numbers written on top of the circles.
- The output must be given in a single line in the form of CIRCLES:x where x is the list of numbers seen on top of the circles

\subsubsection{Sort Circles} 
\textbf{Input Prompt:} \\
- There are several black circles with different sizes.
- The image may also have a single black circle.
- Sort these circles by size, from smallest to largest.
- If there is only a single circles, give the number written on that circle.
- Output must only contain the space separated labels of circles in their sorted order by size.
\begin{figure}[!h]
    \centering
    \frame{\includegraphics[width=\columnwidth]{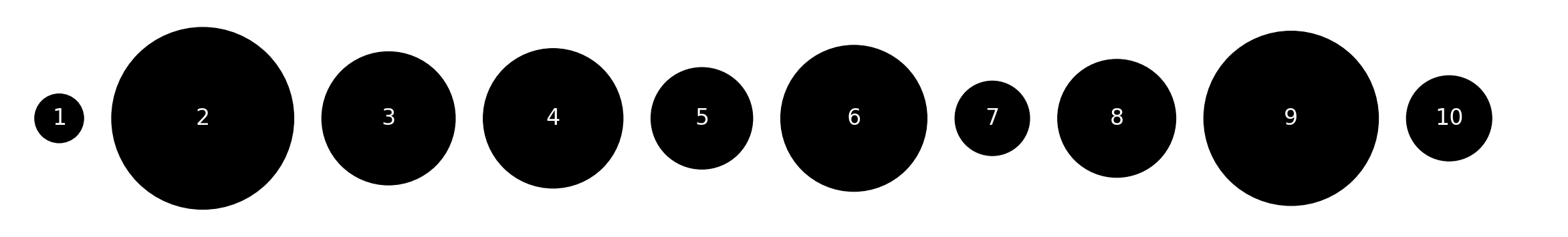}}
    \caption{Sort Circles}
\end{figure}

\subsubsection{Sort Lines} 
\textbf{Input Prompt:} \\
- The image has several parallel lines of varying lengths.
- The image may also contain only a single line.
- Sort these lines by length, from shortest to longest.
- If there is only a single line, give the number of that line.
- Last line of the output must only contain the space separated labels of thee lines in their sorted order by length and nothing else.
\begin{figure}[!h]
    \centering
    \frame{\includegraphics[scale=0.17]{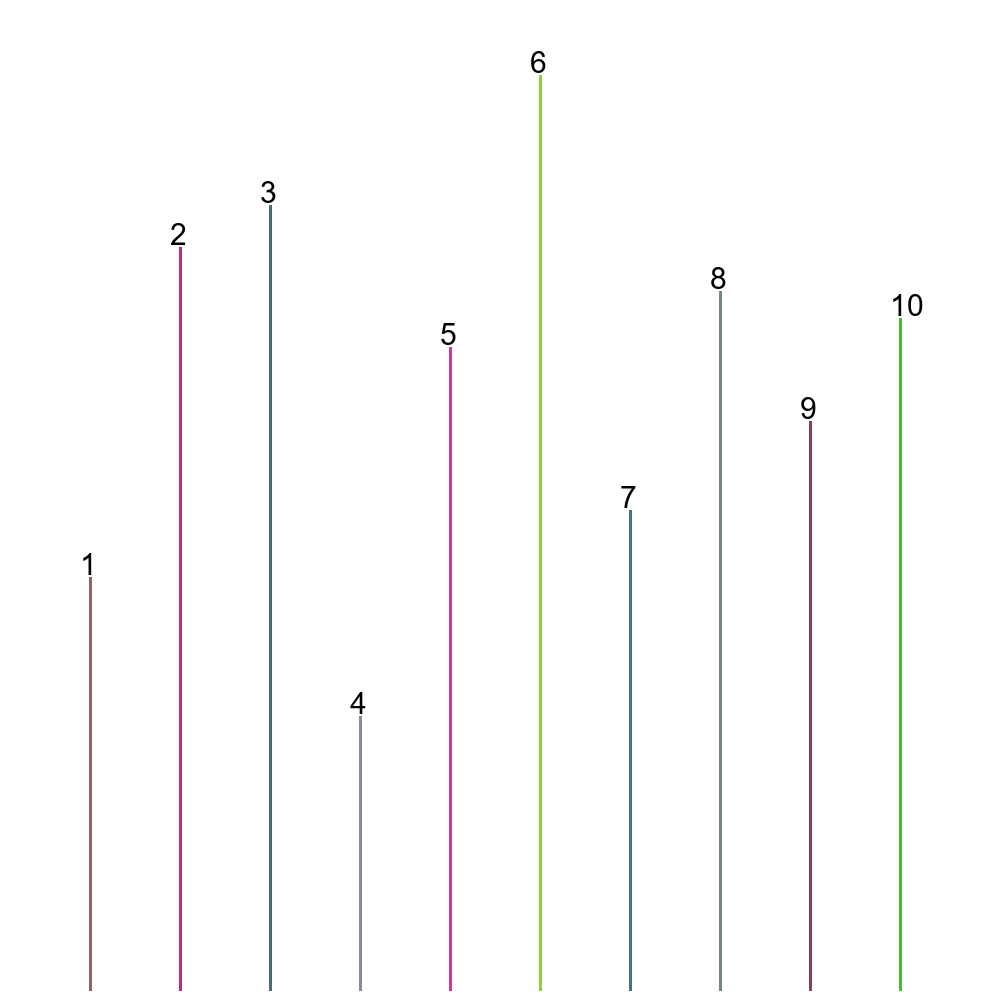}}
    \caption{Sort Lines}
\end{figure}

\subsubsection{Vanishing Objects} 
\textbf{Input Prompt:} \\
- These are two images containing shapes such as circles, triangles and squares
- The images are identical except for the fact that the second image has a few objects missing
- Count the number of missing objects in the second image
- The output must be given in a single line in the form of COUNT:x
\begin{figure}[!h]
    \centering
    % --- LEFT SIDE: TABLE ---
    \begin{minipage}{0.45\columnwidth}
        \centering
        \frame{\includegraphics[scale=0.13]{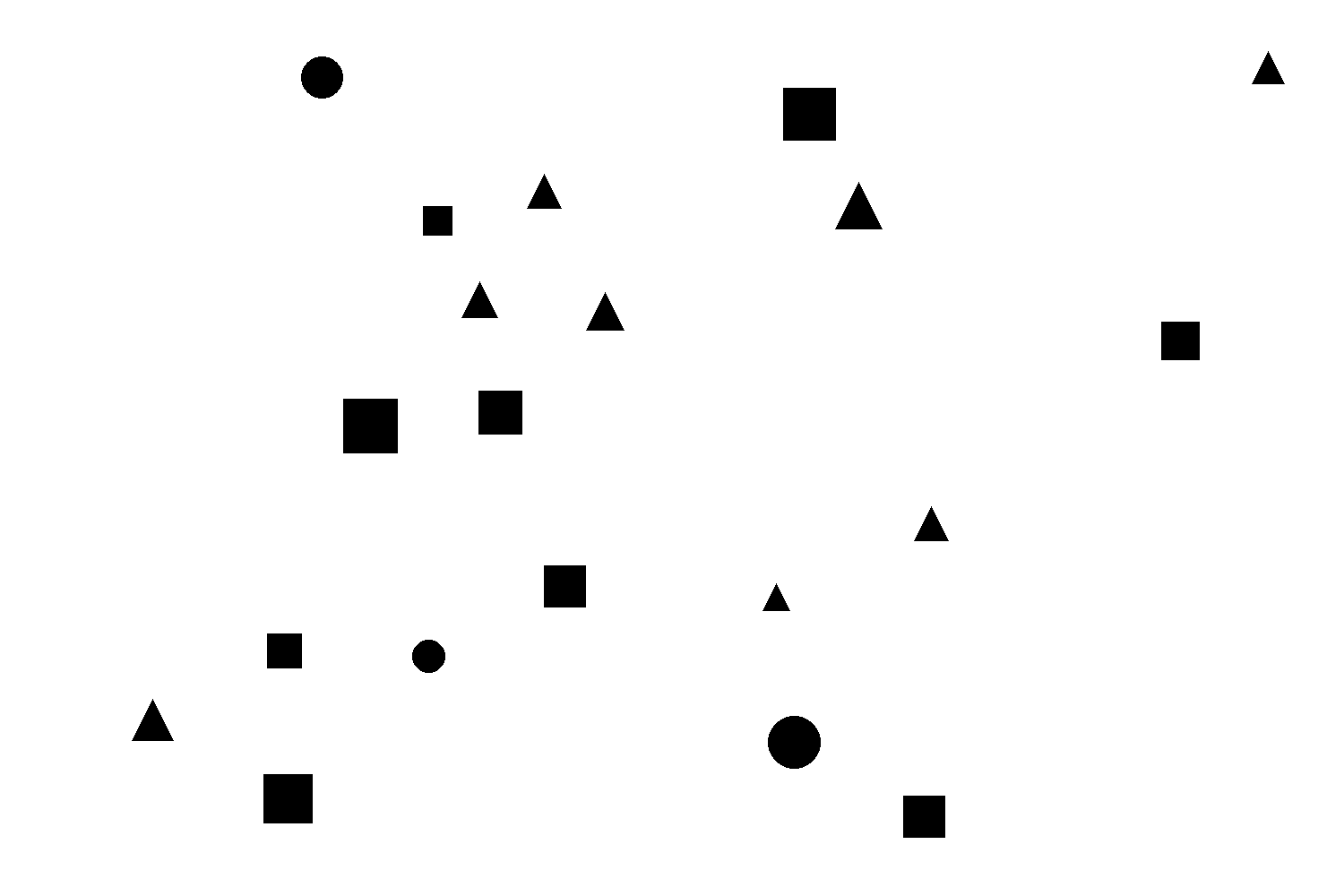}}
        \caption{Vanishing Objects 1st Image}
    \end{minipage}
    \hfill % Adds space between minipages
    % --- RIGHT SIDE: FIGURE ---
    \begin{minipage}{0.45\columnwidth}
        \centering
        \frame{\includegraphics[scale=0.13]{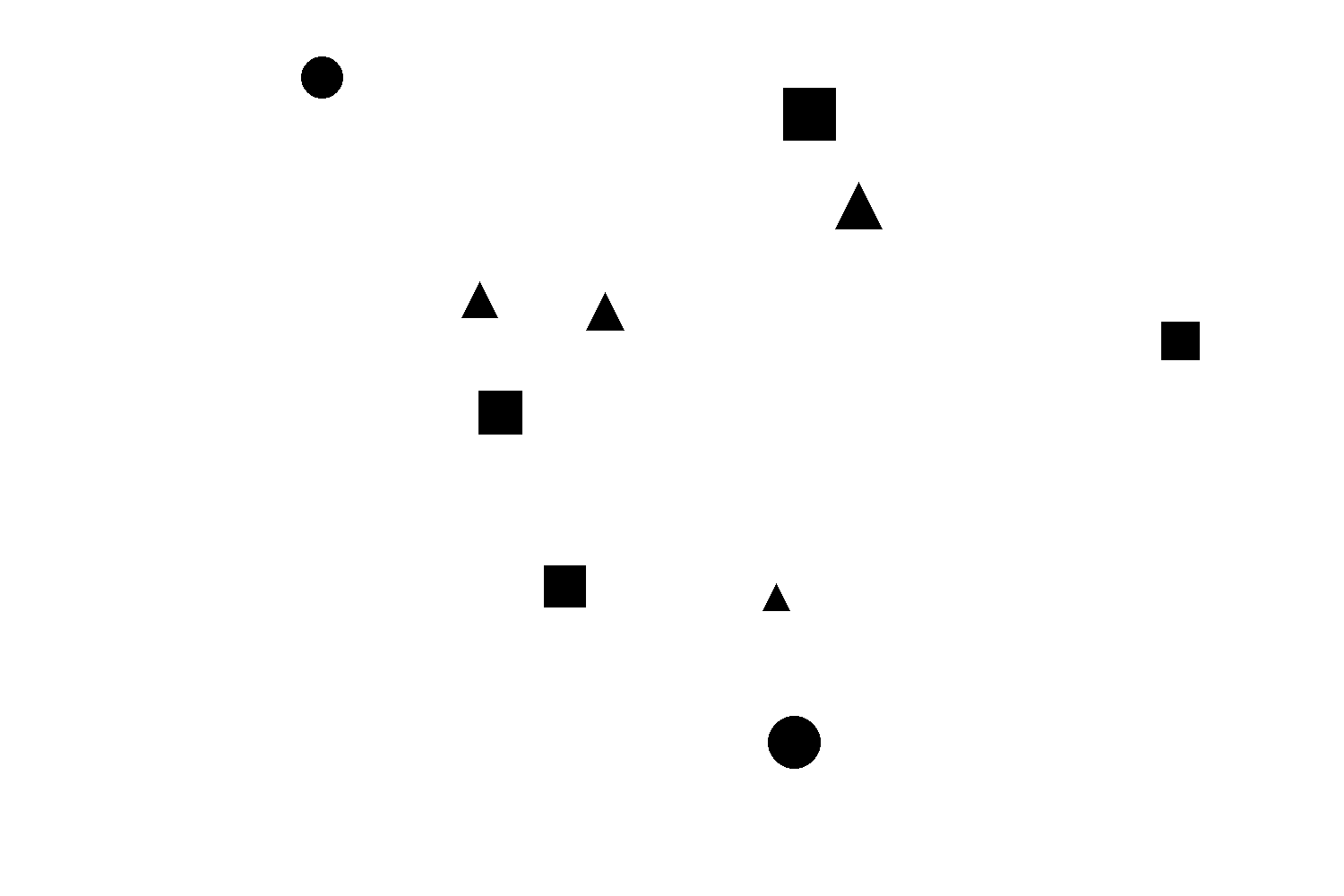}}
        \caption{Vanishing Objects 2nd Image}
    \end{minipage}
\end{figure}

\subsubsection{Water Image} 
\textbf{Input Prompt:} \\
- These are two images containing different shapes in different colours
- The image consists of a horizontal blue line at the centre.
- Some objects of the first image are mirrored along the given horizontal line in the second image.
- Other objects in the second image maintain the same postions as in the first image.
- Determine the number of objects that are mirrored along the horizontal line in the second image.
- The output must be given in a single line in the form of COUNT: x, where x is the number of being mirrored
\begin{figure}[!h]
    \centering
    % --- LEFT SIDE: TABLE ---
    \begin{minipage}{0.45\columnwidth}
        \centering
        \frame{\includegraphics[scale=0.53]{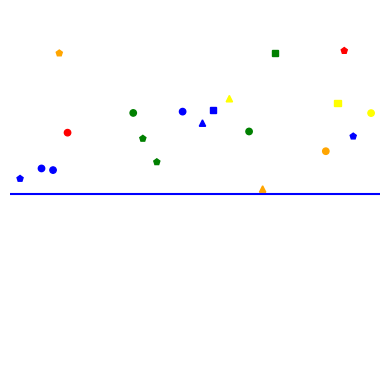}}
        \caption{Water Image First Image}
    \end{minipage}
    \hfill % Adds space between minipages
    % --- RIGHT SIDE: FIGURE ---
    \begin{minipage}{0.45\columnwidth}
        \centering
        \frame{\includegraphics[scale=0.53]{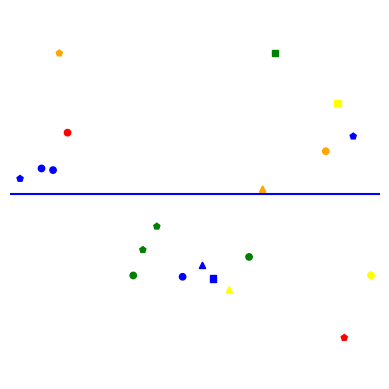}}
        \caption{Water Image Second Image}
    \end{minipage}
\end{figure}

\subsection{Performance on Single Skill Domains}
\label{sec:single_skill_appendix}
Interestingly, we see that the proprietary thinking/reasoning models perform better on 'vsm' as compared to others which indicates that this skill may require higher reasoning capabilities and thus the reasoning models fare better. On the other hand, 'vsr' i.e., Visual Spatial Relationships is an easier skill for models, with almost all of them exhibiting their best performance in it.

\begin{figure*}[!h]
\centering
\includegraphics[width=\textwidth]{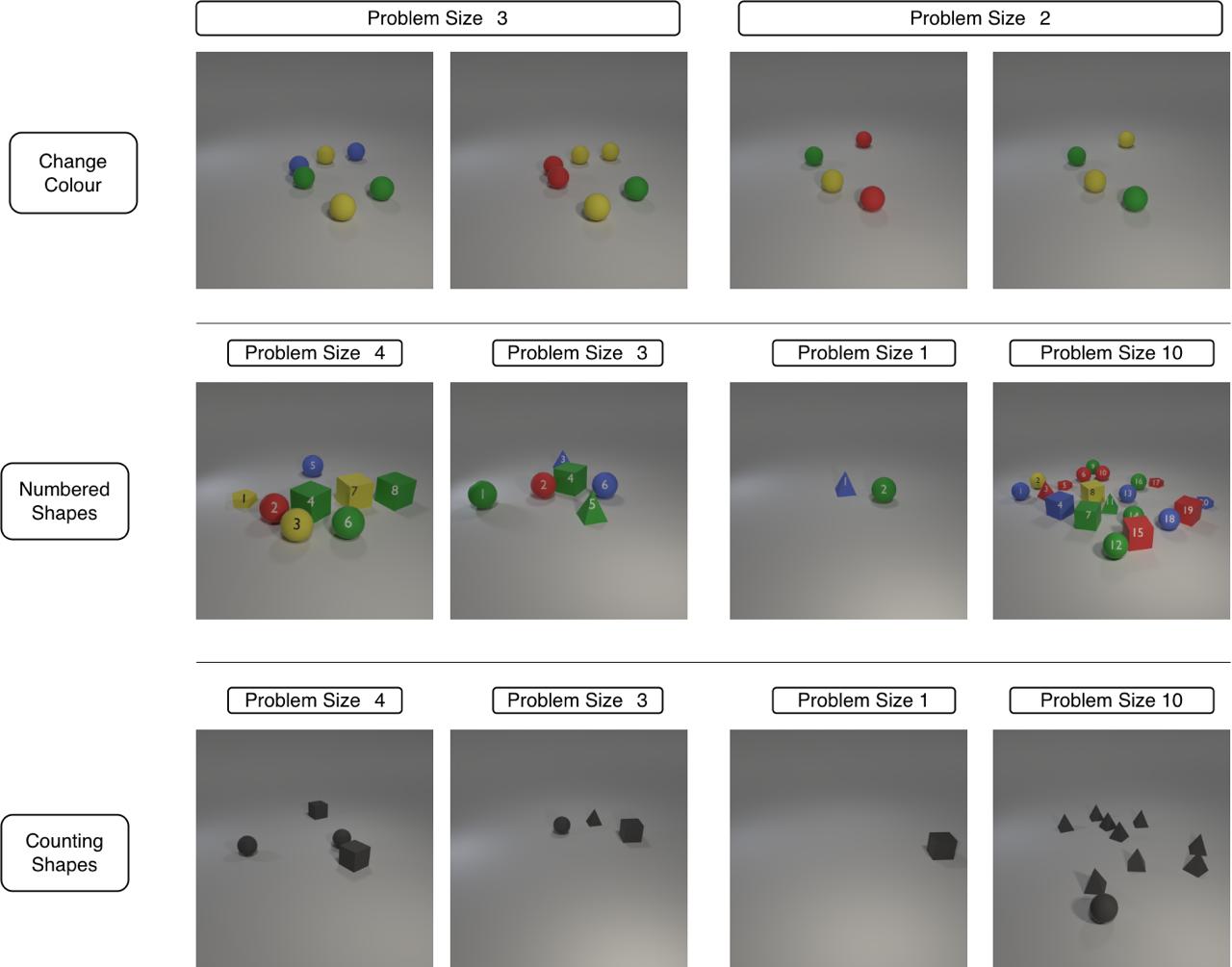}
\caption{Examples from 3D-Percept-V created via Blender for a subset of domains.}
\label{fig:3d_percept_v}
\end{figure*}

\begin{figure*}[!h]
\centering
\includegraphics[width=\textwidth]{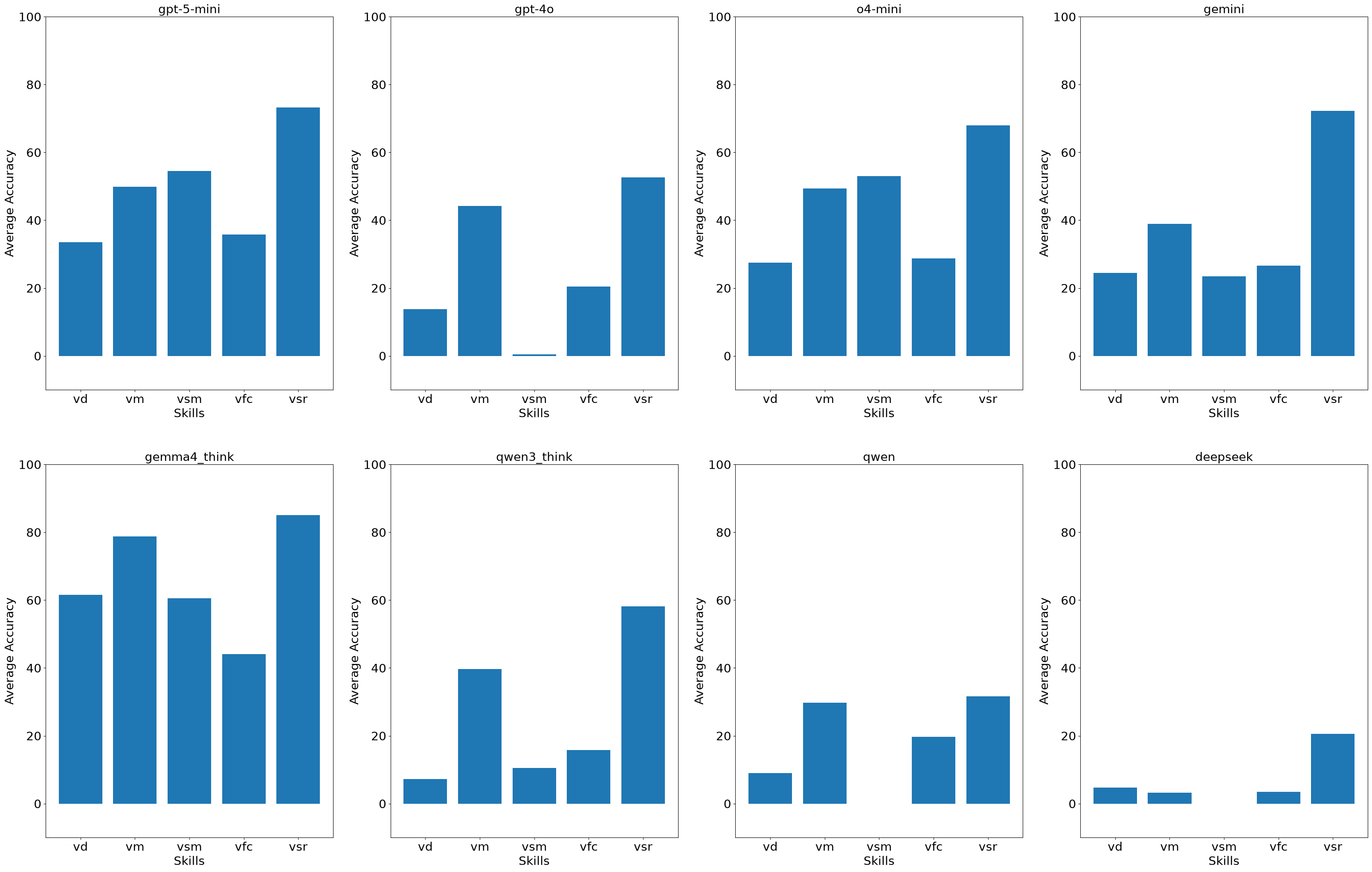}
\caption{The average accuracies of different models on single skill domains categorised by their skills.}
\label{fig:single_skill}
\end{figure*}

\subsection{Overall Performance in Human Study}
Table~\ref{table:human} shows the overall average accuracies of humans and all the evaluated models
\begin{table}[h]
\centering
%\resizebox{\columnwidth}{!}{%
{\small
    \begin{tabular}{|r|c|}
    \hline
    \bf{Model} & \bf{Overall performance}\\
    \hline
    GPT-5-mini & 56.81\%\\
    GPT-4o & 31.94\%\\
    o4-mini & 56.83\%\\
    Gemini & 55.58\%\\
    Qwen-2.5-VL & 14.69\%\\
    Deepseek & 7.08\%\\
    Gemma-4-31B & 68.71\%\\
    Qwen-3-VL & 36.14\%\\
    Human & \textbf{95.43\%}\\
    \hline
    \end{tabular}%
}
\caption{Overall performance in human study}
\label{table:human}
\end{table}

\subsection{One-shot Results}
Table~\ref{table:one_shot_results} presents the results for a subset of the models on Percept-V.

\begin{table*}[!h]
\centering
\small
\begin{tabular}{|l|r|r|r|r||r|}
\hline
{\bf Skill / Domain} & {\bf GPT-5-mini} & {\bf GPT-4o} & {\bf o4-mini} & {\bf Gemini} & {\bf Avg. Acc.}\\
\hline
\multicolumn{6}{|l|}{\bf --- Performance by Skill ---} \\ \hline

Visual Discrimination & 51.46 & 11.17 & 52.51 & 43.57 & 39.68 \\
Visual Memory & 46.55 & 20.93 & 45.96 & 39.38 & 38.2 \\
Visual Sequential Memory & 38.61 & 12.33 & 45.48 & 33.42 & 32.46 \\
Visual Figure Ground & 63.9 & 20.0 & 63.9 & 48.83 & 49.16 \\
Visual Form Constancy & 53.06 & 16.64 & 45.63 & 40.7 & 39.01 \\
Visual Closure & 40.58 & 17.75 & 50.76 & 40.96 & 37.51 \\
Visual Spatial Relationship & 64.61 & 24.5 & 61.59 & 64.11 & 53.7 \\
\hline
Average of skills & 51.25 & 17.62 & 52.26 & 44.42 & 41.39\\
\hline
\hline
\multicolumn{6}{|l|}{\bf --- Performance by Domain ---} \\ \hline
change\_colour & 30.15 & 11.5 & 24.62 & 21.11 & 21.84 \\
circle\_boxes & 26.63 & 21.5 & 28.64 & 24.12 & 25.22 \\
circle\_location & 55.78 & 13.0 & 48.24 & 54.27 & 42.82 \\
circle\_right\_triangle & 100.0 & 59.0 & 100.0 & 96.48 & 88.87 \\
colours\_present & 4.02 & 1.5 & 5.03 & 1.01 & 2.89 \\
comparing\_size & 17.59 & 2.5 & 23.62 & 36.18 & 19.97 \\
count\_coloured\_circles & 41.71 & 16.5 & 35.68 & 46.73 & 35.16 \\
counting\_circles & 57.79 & 23.5 & 48.74 & 53.27 & 45.82 \\
counting\_locations & 35.68 & 9.0 & 35.18 & 28.64 & 27.12 \\
counting\_shapes & 60.8 & 20.2 & 50.75 & 56.28 & 47.01 \\
cross\_and\_knots & 100.0 & 21.0 & 100.0 & 91.46 & 78.11 \\
graph\_counting & 33.17 & 3.0 & 27.14 & 31.16 & 23.62 \\
grid\_path & 51.76 & 0.5 & 46.23 & 0.5 & 24.75 \\
identifying\_shapes & 67.84 & 41.5 & 91.46 & 51.76 & 63.14 \\
inside\_circles & 94.97 & 52.0 & 92.96 & 95.48 & 83.85 \\
layered\_colours & 31.16 & 25.0 & 34.17 & 24.62 & 28.74 \\
layered\_shapes & 21.11 & 12.5 & 19.1 & 19.1 & 17.95 \\
list\_colours & 40.7 & 12.0 & 40.7 & 36.18 & 32.4 \\
list\_shapes & 91.96 & 68.5 & 91.96 & 82.91 & 83.83 \\
locate\_circles\_colour & 99.5 & 1.0 & 98.49 & 63.32 & 65.58 \\
locate\_circles\_shape & 98.9 & 1.0 & 98.99 & 66.83 & 66.45 \\
match\_outline & 58.79 & 17.5 & 79.4 & 71.86 & 56.89 \\
match\_shadow & 51.26 & 16.0 & 70.35 & 48.24 & 46.46 \\
maze\_solving & 48.24 & 15.0 & 31.16 & 46.23 & 35.16 \\
mirror\_image & 30.15 & 17.5 & 27.14 & 20.1 & 23.72 \\
numbered\_shapes & 81.91 & 9.5 & 77.39 & 53.77 & 55.64 \\
sort\_circles & 34.67 & 16.5 & 33.17 & 25.63 & 27.49 \\
sort\_lines & 28.64 & 19.5 & 23.62 & 23.62 & 23.85 \\
vanishing\_objects & 34.17 & 6.0 & 22.11 & 22.61 & 21.22 \\
water\_image & 34.67 & 12.5 & 24.12 & 20.1 & 22.85 \\
\hline
Full Percept-V Dataset & 52.13 & 18.21 & 51.01 & 43.79 & 41.28\\
\hline
\end{tabular}
\caption{MLLM performance across different skills and domains in one-shot setting (in \%).}
\label{table:one_shot_results}
\end{table*}

\begin{figure*}[!h]
\centering
\includegraphics[scale=0.15]{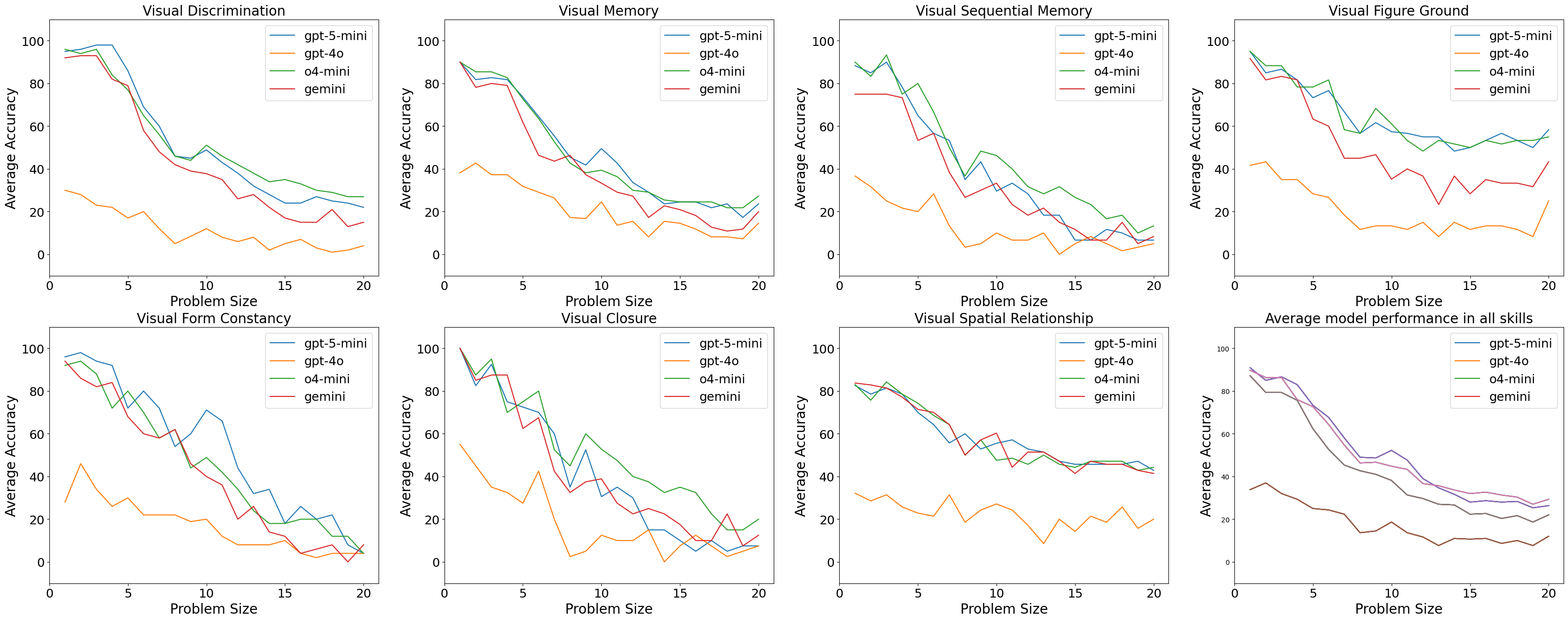}
\caption{The overall accuracy of all models in different skills in one-shot setting.}
\label{fig:size_vs_acc_one_shot}
\end{figure*}

\subsection{Answer Type Result}
\begin{figure*}[!h]
\centering
\includegraphics[width=0.8\textwidth]{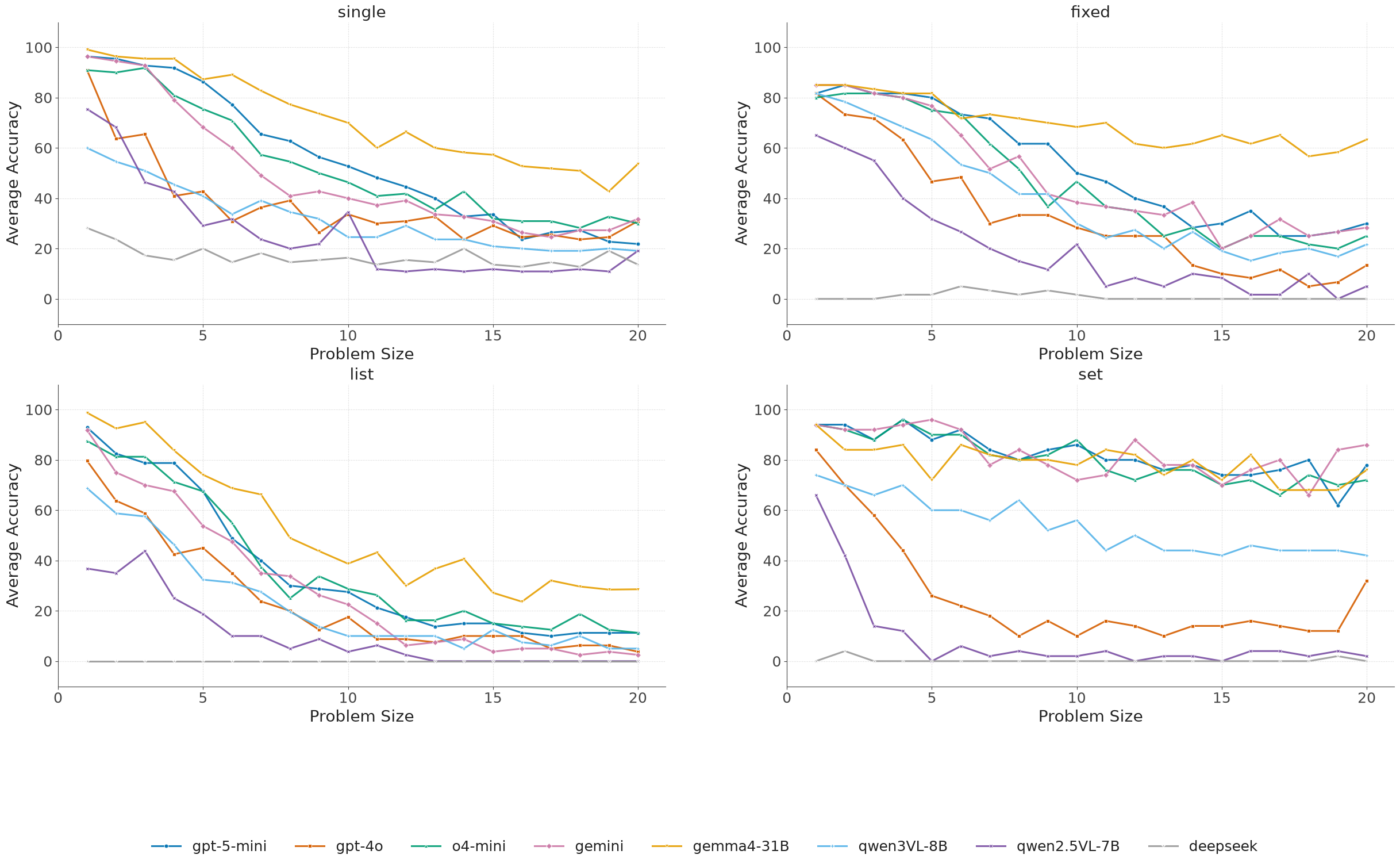}
\caption{The overall accuracy of all models in different answer types.}
\label{fig:answer_type_sizes}
\end{figure*}

An analysis of performance across problem size categorized by answer types is provided in Figure \ref{fig:answer_type_sizes}. Upon inspection, it is found that thinking models like GPT-5-mini, Gemini and o4-mini are relatively robust to size variance in unordered set answer type but the same is not true when it comes to ordered list answer type. This reveals that following an order while answering is a challenge for the evaluated models.

% \subsection{Resolution Results}
% \label{sec:appendix_resolution}
% \begin{table}[h!]
% \centering
% \small
% \begin{adjustbox}{width=1\textwidth}
% \begin{tabular}{|l|r|r|r|r|r|r|}
% \hline
% {\bf Skill / Domain} & {\bf GPT-5-mini Upscale} & {\bf Qwen Upscale} & {\bf GPT-5-mini Original} & {\bf Qwen Original} &  {\bf GPT-5-mini Downscale} & {\bf Qwen Downscale}\\
% \hline
% Visual Discrimination & 54.8 & 11.7 & 55.0 & 11.5 & 45.83 & 10.2 \\
% Visual Memory & 49.94 & 22.83 & 46.67 & 21.44 & 32.92 & 17.22 \\
% Visual Sequential Memory & 42.17 & 4.83 & 43.5 & 4.67 & 29.42 & 4.33 \\
% Visual Figure Ground & 63.12 & 4.88 & 64.75 & 5.75 & 52.0 & 4.62 \\
% Visual Form Constancy & 61.9 & 25.2 & 58.7 & 25.6 & 38.15 & 21.5 \\
% Visual Closure & 45.75 & 5.25 & 44.88 & 5.75 & 32.75 & 5.75 \\
% Visual Spatial Relationship & 71.64 & 30.36 & 66.64 & 27.79 & 57.07 & 24.07 \\
% \hline
% Average of skills & 55.62 & 15.01 & 54.3 & 14.64 & 41.16 & 12.53\\
% \hline
% \end{tabular}
% \end{adjustbox}
% \caption{Comparison of MLLMs' performance across different resolutions (in \%).}
% \label{table:resolution_results1}
% \end{table}

% \subsection{Paraphrasing Results}
% \label{sec:appendix_paraphrasing}
\begin{table}[h!]
\centering
\small
\begin{adjustbox}{width=1\textwidth}
\begin{tabular}{|l|r|r|r|r|}
\hline
{\bf Skill / Domain} & {\bf GPT-5-mini Concise} & {\bf Qwen Concise} & {\bf GPT-5-mini Verbose} & {\bf Qwen Verbose}\\
\hline
Visual Discrimination & 55.3 & 13.91 & 53.55 & 13.45 \\
Visual Memory & 47.55 & 21.87 & 49.27 & 18.77 \\
Visual Sequential Memory & 47.5 & 5.5 & 39.83 & 7.25 \\
Visual Figure Ground & 63.25 & 4.33 & 63.42 & 5.08 \\
Visual Form Constancy & 48.9 & 24.31 & 58.5 & 30.1 \\
Visual Closure & 50.25 & 5.38 & 39.62 & 10.62 \\
Visual Spatial Relationship & 62.64 & 18.07 & 66.29 & 15.5 \\
\hline
Average of skills & 53.63 & 13.34 & 52.93 & 14.4\\
\hline
\end{tabular}
\end{adjustbox}
\caption{Comparison of models' performance across different prompting styles (in \%).}
\label{table:prompt_results}
\end{table}

\begin{table}[h!]
\centering
\small
\begin{tabular}{|l|rr|rr|}
\hline
 & \multicolumn{2}{c|}{\bf GPT-5-mini} & \multicolumn{2}{c|}{\bf Qwen} \\
{\bf Domain} & {\bf 2D} & {\bf 3D} &  {\bf 2D} & {\bf 3D} \\
\hline
change\_colour         & 64.0 & 58.0 & 16.0 & 15.0  \\
colours\_present       & 1.0 &  9.0 & 0.0 &  0.0 \\
count\_coloured\_circles & 90.0 & 78.0  & 62.0 & 62.0  \\
counting\_circles      & 93.0 & 74.0 &  78.0 & 92.0  \\
counting\_shapes       & 95.0 & 94.0  & 67.0 & 81.0  \\
numbered\_shapes       & 98.0 & 19.0 & 16.0 & 17.0 \\
\hline
Average (6 domains)    & 73.50 & 55.33 & 39.8 & 44.5  \\
\hline
\end{tabular}
\caption{Model accuracy (\%) on six Percept-V domains rendered in 2D and in 3D,
matched on problem sizes 1--10 (100 instances per domain in each condition).}
\label{table:3D_results1}
\end{table}

\subsection{Discussion on Visual Memory}
\label{sec:long_term_visual_memory}

In this section we reflect deeper on how "memory" actually functions in humans and theorize the existence of short-term and long-term memory analogues within the MLLM architectures. A standard human (short-term) memory in TVPS-4 works as follows. The human is shown an image, and then the image is removed and a question is asked. The human's goal is to convert the image into a mental representation without access to the question, and answer the question based on this representation. We posit that exactly this analogue happens in an MLLM forward pass. The visual encoder pre-processes the image into a question-independent sequence of latent embeddings (tokens). These representations are held in the context window, which effectively acts as the model's transient short-term memory. The MLLM then uses these tokens to answer the question at hand. Therefore, our experiments essentially test the fidelity and capacity of this short-term representational memory. In this analogy, MLLMs will be considered as using the image (and not their memory) for answering, if the visual representations were computed in a question dependent way, i.e., the model could explicitly use the raw pixels for answering the specific question — this does not happen in a standard MLLM architecture. Conversely, the cognitive definition of "recalling a removed image" can be mapped to an MLLM retrieving visual information stored in its weights during training (long-term memory). To directly test this, we designed a novel experiment to simulate long-term recall. During training phase, we forced the model to encode images into its parametric memory via fine-tuning (Image → Unique Text ID). At test time, we removed the image entirely and queried the model using only the ID (Text ID → Task Answer). Under this cognitive memory constraint, the model's performance collapsed to just 1.23\%.

Together, these results show that MLLMs fail at both maintaining short-term visual representations and retrieving long-term visual memories.

% \AG{
% The argument in Section 3 treats the context window as a short-term memory. The complementary case i.e recall of an image that is no longer available in any form requires the visual content to have been committed to the model's parameters. We test this with the following two-phase protocol. In the encoding phase, we fine-tune Qwen-2.5VL-7B on (image → unique text identifier) pairs drawn from visual-memory tasks in Percept-V, forcing each image to be associated with a distinct symbolic handle. At evaluation time the image is withheld entirely and the model receives only the identifier together with the domain question (identifier → answer), so that any correct response must be retrieved from parametric memory. Accuracy under this constraint falls to 1.23\%. This suggests that the failure lies in retrieval rather than in the association step. MLLMs thus fail on both counts, they neither maintain a faithful short-term visual representation nor form a retrievable long-term one.
% }

\subsection{Finetuning} 
\label{sec:appendix_finetune}
All the models were finetuned using qwen-vl's official codebase. LoRA models were trained with rank 8 and batch-size of 16 on 2xA100 GPUs using $1e-4$ learning rate with cosine-decay. Full-Finetuning runs were trained using 8xA100 GPUs with $2e-6$ for 3 epochs.

% The loss curves for the models  \\ 

% \begin{figure}[h!]
%     \centering
%     \includegraphics[width=\textwidth]{images/image.png}
%     \caption{Training Loss flattens for the models}
%     \label{fig:finet}
% \end{figure}

\subsection{Instructions for Participants}
\label{section:participant_instructions}
Thank you for participating in this experiment! This questionnaire will test your visual perception skills through some simple questions. Please follow the below instructions carefully - 
1. Focus on accuracy and not on speed. There is no time bound , so you can take breaks while solving. However, you have to keep the website open on some tab/window otherwise, your progress will be lost. Also, take sufficient time to solve each question.
2. Please open this questionnaire on laptop/tablet . Some fonts are small, so, you may miss out if you open on smaller devices. Zoom in on images that appear tiny otherwise, this may affect your accuracy!
3. Each question follows a particular answer format. Follow the answer format strictly! Check for spelling mistakes while writing.
4. Try to attempt all the questions. If you do not understand any question, reach out to us!
All the best! 

\subsection{Reproducibility}
\subsubsection{Image Generation}
Image is generated using hyperparameters like --num\_image that give the number of images to be generated per problem size and --num\_size that give the list of problem sizes.
\subsubsection{GPU Hours}
GPU Hours required were around 5 for Deepseek (1B parameters) and Qwen(7B parameters). The number of parameters for o4-mini and GPT-4o are estimated to be 8B and 200B.

\subsection{Limitations}
This paper evaluates MLLMs only against a limited pool of adult college students. Future work may be expanded into including annotators from different age groups and diverse educational background to improve the robustness of the results. Additionally, this paper only identifies the weakness of MLLMs in visual perceptual skills and does not investigate in detail, the rationale behind such weakness.

\end{document}